\documentclass[10pt,twocolumn,letterpaper]{article}
\usepackage[T1]{fontenc}
\usepackage[pagenumbers]{cvpr} %
\ifdefined\XeTeXversion
  \catcode"2013=13
  \def–{\textendash}
\fi

\usepackage{booktabs}
\usepackage{colortbl}
\definecolor{mscdhighlight}{gray}{0.94}
\definecolor{mscdincrease}{HTML}{16763C}
\definecolor{mscddecrease}{HTML}{B32D2E}
\usepackage{multirow}
\usepackage{graphicx}
\usepackage{amsmath}
\usepackage{adjustbox}

\makeatletter
\newcommand{\tableheaderseparator}[2]{%
  \smash{\vrule width\arrayrulewidth
    height\dimexpr\ht\@arstrutbox+\belowrulesep-#1\relax
    depth\dimexpr\dp\@arstrutbox+\aboverulesep+\cmidrulewidth-#2\relax}%
}
\newsavebox{\tableheaderlabelbox}
\newcommand{\tableheaderupperseparator}[1][\ht\@arstrutbox]{%
  \smash{\vrule width\arrayrulewidth height#1
    depth\dimexpr\dp\@arstrutbox+\aboverulesep+\cmidrulewidth\relax}}
\newcommand{\tableheaderlowerseparator}{%
  \tableheaderseparator{0pt}{\dimexpr\aboverulesep+\cmidrulewidth\relax}}
\newcommand{\tableheadersingleseparator}{%
  \tableheaderseparator{\belowrulesep}{\dimexpr\aboverulesep+\cmidrulewidth\relax}}
\makeatother

\definecolor{cvprblue}{rgb}{0.21,0.49,0.74}
\usepackage[pdfversion=1.7,pagebackref,breaklinks,colorlinks,allcolors=cvprblue]{hyperref}
\usepackage[capitalize]{cleveref}

\title{Modality-Shift Contrastive Decoding\\
for Temporally Grounded Audio--Visual Understanding}

\author{
Minyeong Jeong \qquad Yeonju Kim \qquad Yong Man Ro\\
Integrated Vision Language Lab, KAIST\\
{\tt\small \{alsdud6943, yeonju7.kim, ymro\}@kaist.ac.kr}
}

\hypersetup{
    pdftitle={Modality-Shift Contrastive Decoding for Temporally Grounded Audio-Visual Understanding},
    pdfauthor={Minyeong Jeong, Yeonju Kim, Yong Man Ro}
}

\begin{document}
\maketitle
\begin{abstract}
Audio--visual large language models (AV-LLMs) are designed to represent the temporal structure of audio and video, yet it remains unclear how effectively they use this temporal information to relate sounds to visual events.
To examine this, we probe an AV-LLM on an audio--visual synchronization task by shifting the audio so that the queried sound overlaps a different visual event.
We observe that the model still favors the originally aligned event over the newly aligned one, indicating limited sensitivity to changes in audio--visual temporal alignment.
To address this, we propose Modality-Shift Contrastive Decoding (MSCD), a training-free decoding method that emphasizes audio--visual temporal alignment in model predictions.
MSCD contrasts logits under preserved and disrupted temporal alignment to favor predictions supported by aligned audio and video.
We disrupt this alignment by temporally shifting either the audio or the video, while keeping the other modality fixed.
Extensive evaluations with two AV-LLMs across three benchmarks demonstrate the effectiveness of MSCD for temporally grounded audio--visual understanding.
In particular, MSCD improves accuracy over the base models on Daily-Omni by 2.2 and 2.8~percentage points for OmniVinci and video-SALMONN~2+, respectively, suggesting that emphasizing temporal alignment can also benefit broader audio--visual understanding.
\end{abstract}

\section{Introduction}
\label{sec:intro}

\begin{figure}[!t]
    \centering
    \captionsetup[subfigure]{labelformat=simple}
    \renewcommand{\thesubfigure}{(\alph{subfigure})}
    \begin{subfigure}{\linewidth}
        \centering
        \includegraphics[width=\linewidth]{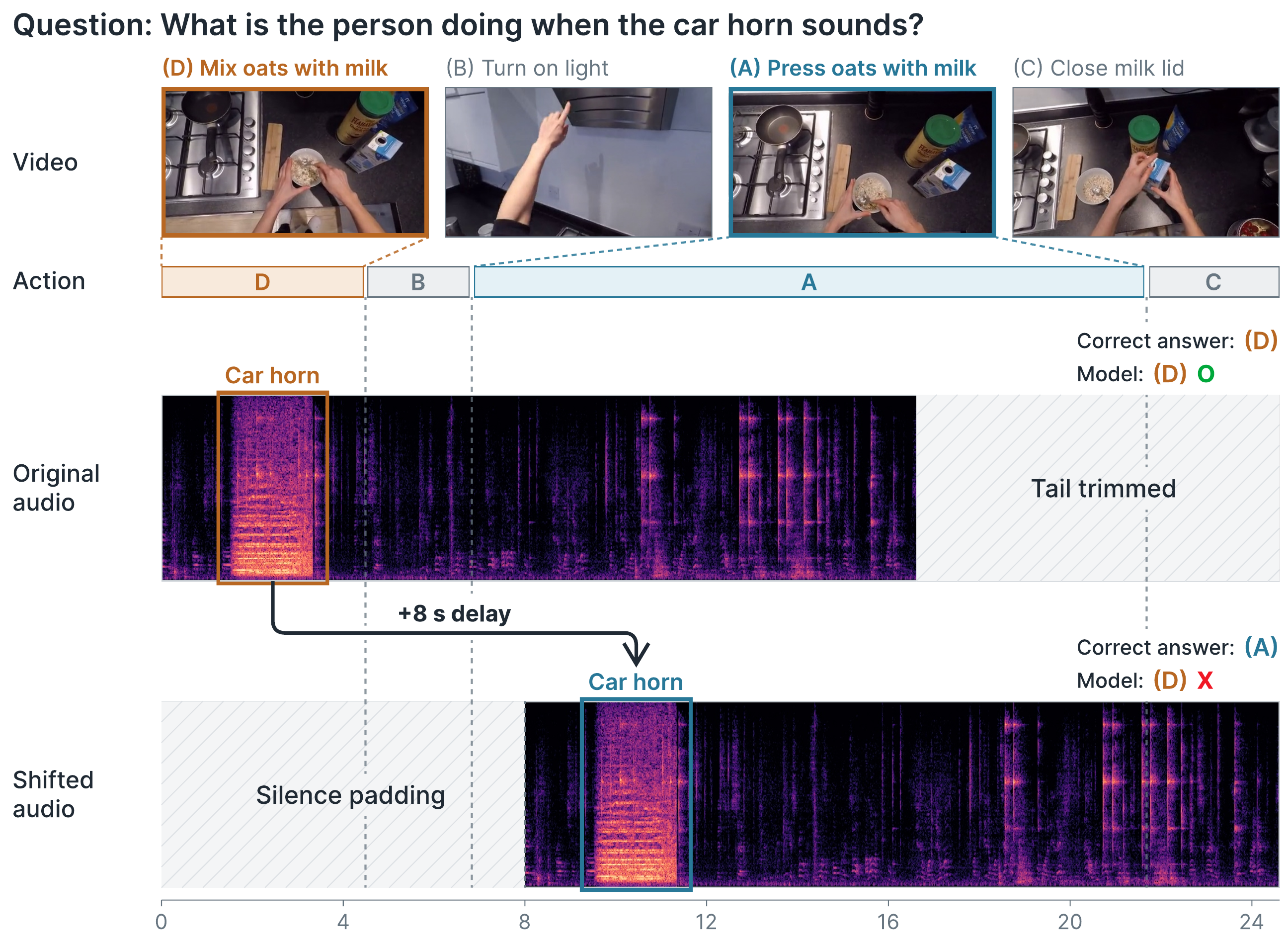}
        \caption{}
        \label{fig:obs_delay_case}
    \end{subfigure}
    \par\bigskip
    \begin{subfigure}{\linewidth}
        \centering
        \includegraphics[width=0.9\linewidth]{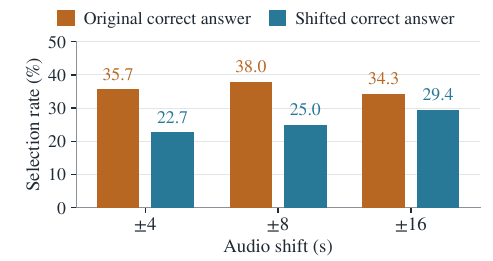}
        \caption{}
        \label{fig:obs_delay_rates}
    \end{subfigure}
    \caption{\textbf{Probing sensitivity to audio--visual temporal alignment through controlled audio shifts on DAVE.}
    \textbf{(a) Specific example.} An 8-second audio delay changes the correct answer from D to A, but the model predicts D in both conditions.
    \textbf{(b) Overall selection rates.} Across all tested shift magnitudes, the baseline model selects the original correct answer more often than the shifted correct answer.}
    \label{fig:obs_delay}
\end{figure}

Understanding real-world situations requires connecting what we see with what we hear.
Audio--visual large language models (AV-LLMs) extend language-based reasoning to visual and auditory inputs, enabling models to jointly process video and audio~\cite{fu2024vita,Qwen2.5-Omni,ICLR2026_5c1863f7,xu2025qwen3,team2026qwen3}.
Yet recognizing events in each modality independently is not sufficient to understand how they relate within a scene.
Models must also associate auditory and visual evidence to interpret the surrounding context and reason about events.
Temporal alignment provides an important cue for establishing these associations: the timing of sounds and visual events helps determine which observations belong together~\cite{NEURIPS2018_c4616f5a,tian2018audio,NEURIPS2025_4d5f03fd,zhou2026dailyomniaudiovisualreasoningtemporal}.

Recent AV-LLMs incorporate mechanisms such as temporal grouping, interleaved modality tokens, and explicit time encodings to represent temporal structure across audio and visual streams~\cite{tang2025video,ICLR2026_5c1863f7,Qwen2.5-Omni}.
However, representing this structure does not necessarily ensure that models use it when relating auditory and visual evidence.
To examine this gap, we conduct a controlled audio-shift probe on DAVE~\cite{NEURIPS2025_4d5f03fd} (\cref{fig:obs_delay}).
We consider questions that ask which visual action occurs when a particular sound is heard, and shift the audio relative to the video so that the sound overlaps with a different action.
If the model effectively associates sounds with visual events based on their relative timing, its answer should change to reflect the newly overlapping event.
However, we observe that the model favors the originally aligned event over the newly aligned one.
\Cref{fig:obs_delay_case} illustrates this behavior: an 8-second audio delay changes the correct answer from D to A, yet the model predicts D in both conditions.
\Cref{fig:obs_delay_rates} shows that this preference persists across all tested shift magnitudes, indicating limited sensitivity to changes in audio--visual temporal alignment.

This observation motivates a decoding strategy that makes model predictions more responsive to audio--visual temporal alignment.
To this end, we propose \emph{Modality-Shift Contrastive Decoding} (MSCD), a training-free decoding method that contrasts predictions under controlled shifts of audio and visual inputs.
To our knowledge, MSCD is the first training-free decoding method specifically designed to strengthen audio--visual temporal alignment in AV-LLMs.
Our key idea is to contrast logits obtained under preserved and disrupted audio--visual temporal alignment.
We disrupt this alignment by cyclically shifting either the encoded audio features or the encoded video features along the temporal axis, while keeping the other modality fixed.
However, cyclically shifting one modality not only changes which sounds and visual events occur together but also alters the temporal order of that modality's own content.
To distinguish these effects, we additionally use an audio--video shift, in which both modalities are shifted by the same amount and therefore remain temporally aligned.
MSCD contrasts logits under the original and audio--video shift conditions with those under the audio shift and video shift conditions.
This contrast favors answers that receive stronger support when audio--visual temporal alignment is preserved, while helping suppress the effects of rearranging each modality's content (\cref{fig:mscd}).

We evaluate MSCD on three audio--visual question-answering benchmarks (DAVE, Daily-Omni, and AVUT)~\cite{NEURIPS2025_4d5f03fd,zhou2026dailyomniaudiovisualreasoningtemporal,yang-etal-2025-audio-centric} using two AV-LLM backbones, OmniVinci and video-SALMONN~2+~\cite{ICLR2026_5c1863f7,tang2025video}.
MSCD improves accuracy over the original models across all three benchmarks, achieving the highest overall accuracy among the compared methods in five of six model--benchmark combinations (\cref{tab:main_w30p2_permodel}).
The improvements are particularly pronounced on DAVE, where MSCD increases accuracy by 5.5 and 3.0~percentage points (pp) for OmniVinci and video-SALMONN~2+, respectively.
Ablation studies further support the proposed contrastive formulation and temporal perturbation strategy, demonstrating the effectiveness of MSCD for temporally grounded audio--visual reasoning.

Our contributions are summarized as follows:
\begin{itemize}
    \item Through a controlled audio-shift probe, we reveal a limitation in an AV-LLM's ability to associate sounds with visual actions based on their temporal correspondence.
    \item We propose Modality-Shift Contrastive Decoding (MSCD). To the best of our knowledge, MSCD is the first training-free decoding method designed to emphasize audio--visual temporal alignment in AV-LLM predictions.
    \item We demonstrate the effectiveness of MSCD for joint audio--visual understanding through evaluations on three benchmarks with two AV-LLM backbones.
\end{itemize}

\section{Related Work}
\label{sec:related}

\paragraph{Cross-Modal Alignment in Audio--Visual LLMs.}
Audio--visual LLMs integrate audio and visual representations with pretrained language models, enabling them to process text, images, audio, and video individually or in combination~\cite{su-etal-2023-pandagpt,panagopoulou2024x,Han_2024_CVPR,zhang-etal-2023-video,cheng2024videollama,pmlr-v235-sun24l,fu2024vita,li2025baichuan,liu2025ola,xu2025qwen3,team2026qwen3}.
To help AV-LLMs understand events using both audio and video, prior work has introduced dedicated modules for cross-modal interaction and alignment.
For example, CAT~\cite{ye2024cat} aggregates question-relevant audio--visual clues, while Meerkat~\cite{chowdhury2024meerkat} supports fine-grained cross-modal grounding.
Recent models also incorporate temporal structure into their representations and reasoning~\cite{dai2026latentomni,ghosh2026audio}.
Qwen2.5-Omni~\cite{Qwen2.5-Omni} and video-SALMONN~2~\cite{tang2025video} interleave audio and visual features according to their temporal correspondence.
OmniVinci~\cite{ICLR2026_5c1863f7} groups the two modalities by time with explicit time embeddings and additionally aligns their representations semantically through contrastive learning.
Beyond architectural design, Thud~\cite{wen2026whenvisionspeaksforsound} diagnoses audio--visual grounding failures through audio shifting, muting, and replacement, and mitigates them through supervised fine-tuning and preference optimization.
SyncRA~\cite{xu2026syncra} diagnoses temporal correspondence failures through audio--visual swaps and introduces a representation-level contrastive objective for fine-tuning AV-LLMs.

Benchmarks such as MUSIC-AVQA~\cite{Li_2022_CVPR}, AVQA~\cite{10.1145/3503161.3548291}, and OmniBench~\cite{NEURIPS2025_2c7c4a12} evaluate audio--visual understanding across a range of tasks.
In particular, WorldSense~\cite{ICLR2026_55dc32df}, Daily-Omni~\cite{zhou2026dailyomniaudiovisualreasoningtemporal}, and AV-Odyssey~\cite{gong-etal-2026-probing} emphasize questions designed to require evidence from both audio and video.
More targeted evaluations assess cross-modal content matching and temporal alignment between the two modalities.
AVUT~\cite{yang-etal-2025-audio-centric} includes tasks that match spoken information to characters, objects, and on-screen text, while TCA-Bench~\cite{10.1007/978-3-032-37490-5_4} evaluates sound--source matching and cross-modal temporal relationships in generated audio--visual captions.
SVHalluc~\cite{Zhang_2026_SVHalluc} diagnoses speech--vision hallucinations by assessing semantic and temporal relationships between spoken information and visual content.
DAVE's Multimodal synchronization task~\cite{NEURIPS2025_4d5f03fd} evaluates temporal alignment by asking which annotated visual action occurs when an added sound is heard.
FAVE~\cite{Lu_2026_CVPR} provides a broader evaluation of audio--visual temporal understanding, covering cross-modal alignment, temporal relationships between events, and moment captioning.

\paragraph{Inference Methods for Multimodal LLMs.}
Training-free inference methods aim to improve how multimodal LLMs use input evidence without updating model parameters.
Contrastive decoding~\cite{li-etal-2023-contrastive} was introduced to guide generation by contrasting predictions from larger and smaller language models.
VCD~\cite{Leng_2024_CVPR} adapts this approach to vision--language models by contrasting predictions from original and distorted visual inputs to mitigate object hallucinations.
Other methods construct contrastive references through temporal perturbations: SEASON~\cite{Wu_2026_CVPR} reduces temporal variation in video features, while TCD~\cite{li-etal-2026-temporal} smooths the audio waveform.

For AV-LLMs, prior work on training-free inference has largely focused on mitigating cross-modal hallucinations.
AVCD~\cite{NEURIPS2025_5b5e70e7} contrasts predictions under attention-guided masking of audio, video, or both.
MAD~\cite{Chung_2026_CVPR} weights modality-specific contrasts according to their relevance to the question, and CAD~\cite{deng2026cad} shifts weight from the joint prediction to unimodal predictions when the modalities conflict.
RED~\cite{ra2026relevant} identifies the evidence a question requires and strengthens its contribution.
Beyond contrastive decoding, other methods intervene inside the decoder.
ASD~\cite{pmlr-v306-jung26c} mitigates hallucinations by increasing attention to cross-modal sink tokens that encode integrated audio--visual information.
FMD~\cite{jung-etal-2026-fork} reduces modality bias by processing audio and video inputs separately in early decoder layers and then merging their hidden states.

While model architectures and benchmarks increasingly address cross-modal alignment to support joint audio--visual understanding, training-free decoding methods that target audio--visual temporal alignment remain underexplored.
We therefore present MSCD, a contrastive decoding framework that emphasizes this alignment during generation.

\section{Method}
\label{sec:method}

\subsection{Preliminaries}
\label{sec:preliminaries}

\paragraph{Audio--visual input representations.}
Audio--visual large language models extract features with modality-specific encoders, such as Whisper-based audio encoders~\cite{pmlr-v202-radford23a} and SigLIP vision encoders~\cite{Zhai_2023_ICCV}.
These features are then projected into the language model's embedding space~\cite{tang2025video,ICLR2026_5c1863f7}.
We denote the encoded audio and video feature sequences by $A$ and $V$, respectively.
In a temporally interleaved representation, visual and audio tokens corresponding to the same time interval form a temporal group.
For $K$ such groups, let $V_k$ and $A_k$ denote the visual and audio token blocks in group $k$.
We write
\begin{equation}
\begin{aligned}
    V_k &= [\mathbf v_{k,1},\ldots,\mathbf v_{k,n_k^v}],\\
    A_k &= [\mathbf a_{k,1},\ldots,\mathbf a_{k,n_k^a}],
\end{aligned}
\label{eq:modality_blocks}
\end{equation}
where $\mathbf v_{k,i}$ and $\mathbf a_{k,j}$ are individual visual and audio token embeddings, and $n_k^v$ and $n_k^a$ are their respective counts.
The visual and audio token blocks are interleaved as
\begin{equation}
    X_{\rm AV}=[V_1,A_1,\ldots,V_K,A_K].
    \label{eq:interleaved_input}
\end{equation}
These tokens are combined with a text-token sequence $Q$ to form the input to the language model.
The backbone encodes token positions and timestamps through its native positional and temporal mechanisms, including rotary position embeddings~\cite{SU2024127063} and model-specific temporal extensions~\cite{tang2025video,ICLR2026_5c1863f7}.

\paragraph{Contrastive decoding.}
We consider input-conditioned contrastive decoding, which contrasts logits from the same model under original and perturbed inputs~\cite{Leng_2024_CVPR,NEURIPS2025_5b5e70e7}.
At decoding step $t$, a model with parameters $\theta$ predicts the next token $y_t$ conditioned on $A$, $V$, $Q$, and the previously generated tokens $y_{<t}=[y_1,\ldots,y_{t-1}]$.
Let $f_\theta$ denote the model that maps these inputs to the vocabulary logits $\mathbf z_t$.
The next-token distribution is
\begin{equation}
\begin{aligned}
    \mathbf z_t &= f_\theta(A,V,Q,y_{<t}),\\
    p_\theta(\cdot\mid A,V,Q,y_{<t})
        &= \operatorname{softmax}(\mathbf z_t).
\end{aligned}
\label{eq:base_prediction}
\end{equation}

Let $\mathbf z'_t$ denote the logits from the same model under a perturbed reference input, with $Q$ and $y_{<t}$ fixed.
Contrastive decoding defines
\begin{equation}
\begin{aligned}
    \mathbf z_t^{\mathrm{CD}}
        &= \mathbf z_t+\alpha\bigl(\mathbf z_t-\mathbf z'_t\bigr),\\
    p_{\rm CD}(\cdot\mid A,V,Q,y_{<t})
        &= \operatorname{softmax}(\mathbf z_t^{\mathrm{CD}}),
\end{aligned}
    \label{eq:contrastive_decoding}
\end{equation}
where $\alpha\geq0$ controls the contrastive strength.
The contrastive term increases the relative preference for tokens with larger logit differences between the original and perturbed inputs.

\begin{figure*}[t]
    \centering
    \includegraphics[width=\textwidth]{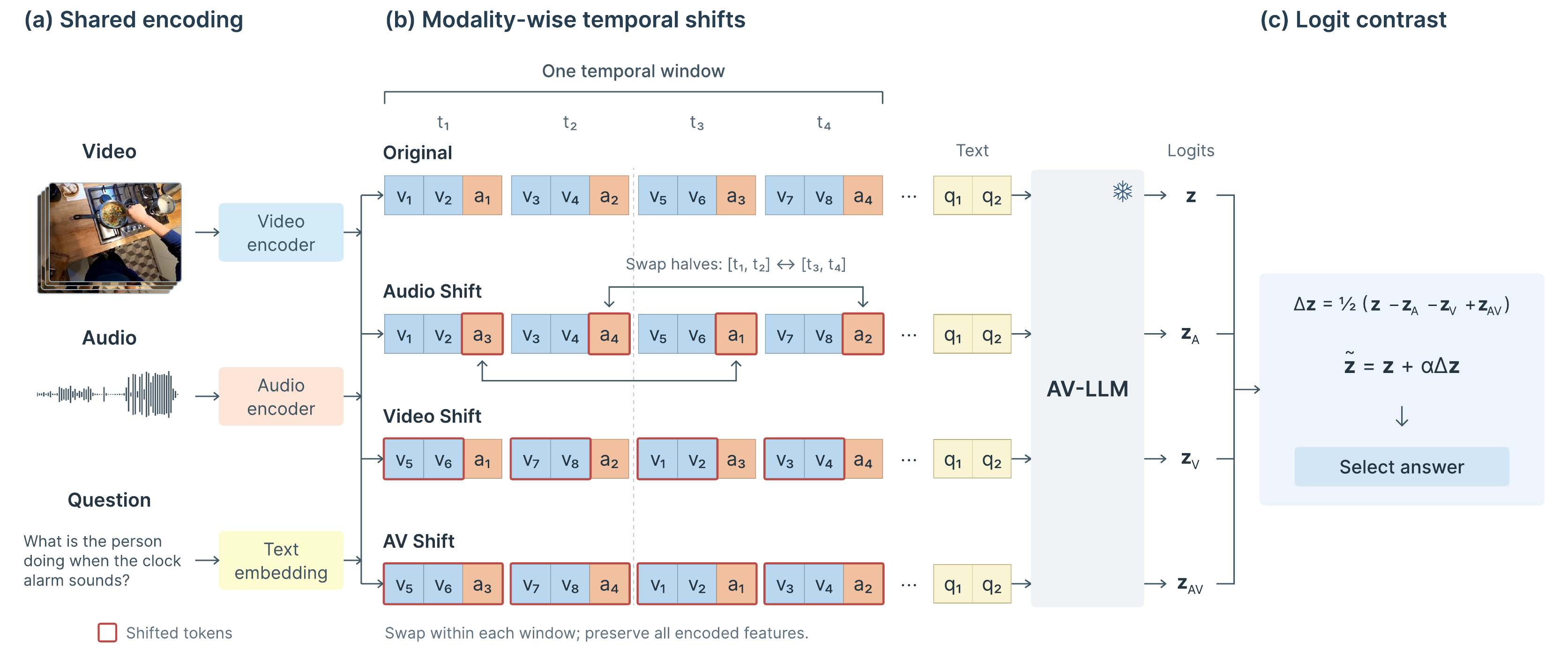}
    \caption{Overview of Modality-Shift Contrastive Decoding (MSCD).
    MSCD swaps the two temporal halves of encoded features within each window to construct audio-shifted, video-shifted, and audio--video-shifted inputs.
    Logits from the original and shifted inputs are contrasted to emphasize audio--visual temporal alignment during decoding.}
    \label{fig:mscd}
\end{figure*}

\subsection{Motivating Observation}
\label{sec:analysis}
\label{sec:obs_delay}

To examine AV-LLMs' sensitivity to changes in audio--visual temporal alignment, we conduct a controlled audio-shift probe on OmniVinci using the \emph{Multimodal synchronization} task from DAVE~\cite{NEURIPS2025_4d5f03fd}.
Each clip contains four annotated actions, and an external cue sound is added to its audio track.
The question asks which visual action occurs when the cue sound is heard.
If the model can associate sounds with visual actions based on their relative timing, shifting the audio so that the cue sound overlaps a different visual action should lead the model to select that action.

We test this by shifting the entire mixed audio waveform by $\delta\in\{\pm4,\pm8,\pm16\}$ seconds, keeping the video, question, and answer options fixed.
Positive shifts delay the audio, whereas negative shifts advance it.
We use silence padding and trimming to preserve the original audio duration.
Using the sound annotations in DAVE, we define the correct answer after the shift as the action with the largest temporal overlap with the cue sound.

We then evaluate examples in which the audio shift changes the correct answer to a different action.
We refer to the correct answers before and after the audio shift as the \emph{original} and \emph{shifted correct answers}, respectively.
For each shift magnitude, we pool eligible examples from both directions and measure how often each answer is selected on the shifted inputs.
The selection criteria and full protocol are provided in \cref{sec:supp_delay}.

Across the tested shift magnitudes, the model selects the original correct answer in 34.3--38.0\% of cases, compared with 22.7--29.4\% for the shifted correct answer (\cref{fig:obs_delay_rates}).
The remaining predictions select other actions or \emph{none of the above}.
This preference indicates that the model's predictions do not consistently reflect the changed audio--visual temporal alignment.
Access to both modalities therefore does not ensure that the model effectively uses their temporal relationship to select an answer.
These findings motivate a decoding strategy that makes temporal alignment a more decisive factor in the model's predictions.

\subsection{Modality-Shift Contrastive Decoding}
\label{sec:mscd}

In this section, we introduce MSCD, a contrastive decoding method for temporally grounded audio--visual understanding.
MSCD contrasts logits from the original, audio shift, video shift, and audio--video shift conditions to emphasize audio--visual temporal alignment.
\Cref{fig:mscd} illustrates the overall framework of MSCD.

\subsubsection{Modality-Shift Perturbations}
\label{sec:perturbations}

We divide the video timeline into consecutive, non-overlapping time intervals with a duration of $w$ seconds, which we refer to as temporal windows.
Within each window, we implement temporal shifts by swapping the first and second halves of the corresponding encoded feature sequences.
This cyclic shift preserves the order within each half and retains all encoded features.
Within each window, audio features are shifted along their temporal sequence, whereas visual tokens are shifted as complete temporal blocks to preserve their spatial organization.

Let $\mathbf h=[\mathbf h_0,\ldots,\mathbf h_{m-1}]$ denote a sequence of $m$ audio feature vectors or visual temporal blocks within a single window.
We set the target temporal displacement to half the window duration, giving $w/2$ seconds for each full window independently of the total video duration.
We determine how many audio feature vectors or visual temporal blocks correspond to the target time shift, and round this number to the nearest integer $s$.
The cyclic-shift operator $R_s$ moves the last $s$ elements to the front while preserving the order within both parts:
\begin{equation}
    R_s\mathbf h=
    \big[\mathbf h_{m-s},\ldots,\mathbf h_{m-1},
    \mathbf h_0,\ldots,\mathbf h_{m-s-1}\big].
    \label{eq:cyclic_shift}
\end{equation}

We apply this operation independently within each window of $A$ and $V$, denoting the resulting operations by $R_A$ and $R_V$, respectively.
An audio shift applies $R_A$ to $A$ while keeping $V$ fixed, and a video shift applies $R_V$ to $V$ while keeping $A$ fixed.
An audio--video shift applies both operators with the same target temporal displacement to preserve audio--visual temporal alignment.
These three conditions are illustrated in \cref{fig:mscd}(b), and their logits are as follows:
\begin{equation}
\begin{aligned}
    \mathbf z_A &= f_\theta(R_A A,\hphantom{R_V}V,Q,y_{<t}),\\
    \mathbf z_V &= f_\theta(\hphantom{R_A}A,R_V V,Q,y_{<t}),\\
    \mathbf z_{AV} &= f_\theta(R_A A,R_V V,Q,y_{<t}).
\end{aligned}
\label{eq:shifted_logits}
\end{equation}
Here, the text-token sequence $Q$ and generated prefix $y_{<t}$ are fixed across all conditions.
For brevity, we omit the decoding-step index $t$ from the logit notation.
MSCD requires four decoder evaluations per decoding step while sharing media encoding across the four conditions.
Backbone-specific implementation details are provided in \cref{sec:supp_shifts}.

\subsubsection{Contrastive Logit Combination}
\label{sec:logit_combination}

\begin{table*}[t]
\centering
\small
\setlength{\tabcolsep}{4pt}
\caption{Main results (accuracy, \%) on DAVE Multimodal synchronization, Daily-Omni, and AVUT. Daily-Omni subtasks, left to right: AV Event Alignment, Comparative, Context Understanding, Event Sequence, Inference, Reasoning; AVUT: Audio-Visual Character Matching, Audio-Visual Object Matching, Audio-Visual Text Matching. Avg. denotes accuracy over all evaluated questions in each benchmark. Subscripts indicate changes from the corresponding Base in pp. Bold marks the highest accuracy within each backbone and column.}
\label{tab:main_w30p2_permodel}
\begin{adjustbox}{max width=\linewidth}
\renewcommand{\arraystretch}{1.12}
\begin{tabular}{@{}l|c|cccccc|c|ccc|c@{}}
\toprule
\multicolumn{1}{@{}l!{\tableheaderupperseparator}}{\multirow{2}{*}{Method}} & \multicolumn{1}{c!{\tableheaderupperseparator}}{\multirow{2}{*}{DAVE}} & \multicolumn{7}{c!{\tableheaderupperseparator}}{Daily-Omni} & \multicolumn{4}{c}{AVUT} \\
\cmidrule(lr){3-9} \cmidrule(lr){10-13}
\multicolumn{1}{@{}l!{\tableheaderlowerseparator}}{} & \multicolumn{1}{c!{\tableheaderlowerseparator}}{} & AV Align. & Comp. & Ctx. Und. & Evt. Seq. & Infer. & \multicolumn{1}{c!{\tableheadersingleseparator}}{Reas.} & \multicolumn{1}{c!{\tableheaderlowerseparator}}{Avg.} & AVCM & AVOM & \multicolumn{1}{c!{\tableheadersingleseparator}}{AVTM} & Avg. \\
\midrule
\multicolumn{13}{@{}l}{\textit{OmniVinci (9B)}} \\
\arrayrulecolor{black!70}
\midrule[\cmidrulewidth]
\arrayrulecolor{black}
Base & 44.3 & 48.3 & \textbf{73.3} & 57.5 & 54.2 & 76.0 & 72.6 & 61.2 & \textbf{67.1} & 63.0 & 70.1 & 66.5 \\
+ AVCD~\cite{NEURIPS2025_5b5e70e7} & 43.6 & 47.9 & 71.8 & 56.5 & 54.2 & 74.7 & 72.6 & 60.6 & 66.7 & 62.8 & 70.4 & 66.3 \\
+ ASD~\cite{pmlr-v306-jung26c} & 44.3 & 48.3 & \textbf{73.3} & 57.5 & 54.2 & 76.0 & 72.6 & 61.2 & \textbf{67.1} & 63.0 & 70.1 & 66.5 \\
+ MAD~\cite{Chung_2026_CVPR} & 43.2 & 46.2 & 72.5 & 56.0 & 52.0 & \textbf{76.6} & 71.4 & 59.7 & 65.5 & 61.5 & \textbf{71.7} & 65.8 \\
+ CAD~\cite{deng2026cad} & 43.2 & 45.8 & 72.5 & 56.5 & 51.6 & 76.0 & 70.9 & 59.5 & 65.5 & 61.0 & \textbf{71.7} & 65.6 \\
\rowcolor{mscdhighlight}
\textbf{+ MSCD} & \textbf{49.8}\textsubscript{\textcolor{mscdincrease}{(+5.5)}} & \textbf{54.6}\textsubscript{\textcolor{mscdincrease}{(+6.3)}} & \textbf{73.3}\textsubscript{(+0.0)} & \textbf{61.7}\textsubscript{\textcolor{mscdincrease}{(+4.1)}} & \textbf{54.9}\textsubscript{\textcolor{mscdincrease}{(+0.7)}} & 76.0\textsubscript{(+0.0)} & \textbf{73.1}\textsubscript{\textcolor{mscdincrease}{(+0.6)}} & \textbf{63.3}\textsubscript{\textcolor{mscdincrease}{(+2.2)}} & \textbf{67.1}\textsubscript{(+0.0)} & \textbf{65.3}\textsubscript{\textcolor{mscdincrease}{(+2.3)}} & 71.4\textsubscript{\textcolor{mscdincrease}{(+1.3)}} & \textbf{67.7}\textsubscript{\textcolor{mscdincrease}{(+1.2)}} \\
\midrule
\multicolumn{13}{@{}l}{\textit{video-SALMONN 2+ (7B)}} \\
\arrayrulecolor{black!70}
\midrule[\cmidrulewidth]
\arrayrulecolor{black}
Base & 42.5 & 50.4 & 64.9 & 55.4 & 55.9 & 79.9 & 74.9 & 61.6 & 60.0 & 56.4 & 71.1 & 61.8 \\
+ AVCD~\cite{NEURIPS2025_5b5e70e7} & 42.0 & 50.0 & 64.1 & 56.0 & 55.9 & 79.9 & 73.7 & 61.3 & 60.0 & 56.1 & 70.7 & 61.6 \\
+ ASD~\cite{pmlr-v306-jung26c} & 42.5 & 50.4 & 64.9 & 55.4 & 55.9 & 79.9 & 74.9 & 61.6 & 60.0 & 56.4 & 71.1 & 61.8 \\
+ MAD~\cite{Chung_2026_CVPR} & 42.6 & 48.7 & 64.1 & 57.0 & 55.6 & 77.9 & 75.4 & 61.2 & \textbf{60.7} & \textbf{58.4} & 71.1 & \textbf{62.8} \\
+ CAD~\cite{deng2026cad} & 42.7 & 48.7 & 64.1 & 57.0 & 55.6 & 78.6 & 75.4 & 61.2 & 60.4 & 58.2 & 70.7 & 62.5 \\
\rowcolor{mscdhighlight}
\textbf{+ MSCD} & \textbf{45.5}\textsubscript{\textcolor{mscdincrease}{(+3.0)}} & \textbf{53.4}\textsubscript{\textcolor{mscdincrease}{(+2.9)}} & \textbf{67.2}\textsubscript{\textcolor{mscdincrease}{(+2.3)}} & \textbf{59.6}\textsubscript{\textcolor{mscdincrease}{(+4.1)}} & \textbf{59.5}\textsubscript{\textcolor{mscdincrease}{(+3.6)}} & \textbf{81.2}\textsubscript{\textcolor{mscdincrease}{(+1.3)}} & \textbf{76.6}\textsubscript{\textcolor{mscdincrease}{(+1.7)}} & \textbf{64.4}\textsubscript{\textcolor{mscdincrease}{(+2.8)}} & 60.0\textsubscript{(+0.0)} & 57.9\textsubscript{\textcolor{mscdincrease}{(+1.5)}} & \textbf{71.4}\textsubscript{\textcolor{mscdincrease}{(+0.3)}} & 62.4\textsubscript{\textcolor{mscdincrease}{(+0.6)}} \\
\bottomrule
\end{tabular}
\end{adjustbox}
\end{table*}

\setcounter{topnumber}{1}
\begin{table*}[t]
\centering
\small
\setlength{\tabcolsep}{4pt}
\caption{Ablation study on the contrastive formulation (accuracy, \%). $\mathbf z$, $\mathbf z_A$, $\mathbf z_V$, and $\mathbf z_{AV}$ denote logits from the original, audio shift, video shift, and audio--video shift conditions, respectively. Incorporating the audio--video shift yields higher accuracy than either single-shift contrast or their mean. Subscripts show changes from Base in pp, and shading marks the default setting.}
\label{tab:abl_contrast}
\begin{adjustbox}{max width=\linewidth}
\renewcommand{\arraystretch}{1.15}
\begin{tabular}{@{}ll|ccc|ccc@{}}
\toprule
\multirow{2}{*}{Variant} & \multicolumn{1}{l!{\tableheaderupperseparator}}{\multirow{2}{*}{Contrastive term $\Delta\mathbf z$}} & \multicolumn{3}{c}{OmniVinci} & \multicolumn{3}{c}{video-SALMONN 2+} \\
\cmidrule(lr){3-5} \cmidrule(lr){6-8}
 & \multicolumn{1}{l!{\tableheaderlowerseparator}}{} & DAVE & Daily-Omni & \multicolumn{1}{c!{\tableheadersingleseparator}}{AVUT} & DAVE & Daily-Omni & AVUT \\
\midrule
Base & & 44.3 & 61.2 & 66.5 & 42.5 & 61.6 & 61.8 \\
Audio shift & $\mathbf z-\mathbf z_A$ & 48.6\textsubscript{\textcolor{mscdincrease}{(+4.3)}} & 58.1\textsubscript{\textcolor{mscddecrease}{(-3.0)}} & 62.3\textsubscript{\textcolor{mscddecrease}{(-4.2)}} & 44.6\textsubscript{\textcolor{mscdincrease}{(+2.1)}} & 61.6\textsubscript{(+0.0)} & 61.6\textsubscript{\textcolor{mscddecrease}{(-0.2)}} \\
Video shift & $\mathbf z-\mathbf z_V$ & 38.0\textsubscript{\textcolor{mscddecrease}{(-6.2)}} & 55.3\textsubscript{\textcolor{mscddecrease}{(-5.8)}} & 57.1\textsubscript{\textcolor{mscddecrease}{(-9.5)}} & 41.1\textsubscript{\textcolor{mscddecrease}{(-1.4)}} & 62.5\textsubscript{\textcolor{mscdincrease}{(+0.9)}} & 60.5\textsubscript{\textcolor{mscddecrease}{(-1.2)}} \\
Mean & $\mathbf z-\tfrac12(\mathbf z_A+\mathbf z_V)$ & 43.3\textsubscript{\textcolor{mscddecrease}{(-0.9)}} & 58.1\textsubscript{\textcolor{mscddecrease}{(-3.1)}} & 60.5\textsubscript{\textcolor{mscddecrease}{(-6.0)}} & 43.4\textsubscript{\textcolor{mscdincrease}{(+0.9)}} & 63.0\textsubscript{\textcolor{mscdincrease}{(+1.4)}} & 61.8\textsubscript{(+0.0)} \\
\rowcolor{mscdhighlight}
Ours & $\tfrac12(\mathbf z-\mathbf z_A-\mathbf z_V+\mathbf z_{AV})$ & \textbf{49.8}\textsubscript{\textcolor{mscdincrease}{(+5.5)}} & \textbf{63.3}\textsubscript{\textcolor{mscdincrease}{(+2.2)}} & \textbf{67.7}\textsubscript{\textcolor{mscdincrease}{(+1.2)}} & \textbf{45.5}\textsubscript{\textcolor{mscdincrease}{(+3.0)}} & \textbf{64.4}\textsubscript{\textcolor{mscdincrease}{(+2.8)}} & \textbf{62.4}\textsubscript{\textcolor{mscdincrease}{(+0.6)}} \\
\bottomrule
\end{tabular}
\end{adjustbox}
\end{table*}

\begin{table}[t]
\centering
\small
\setlength{\tabcolsep}{4pt}
\caption{Ablation study on temporal perturbation strategies (accuracy, \%). Cyclic shift outperforms positional offset and random permutation. Subscripts show changes from Base in pp, and shading marks the default setting.}
\label{tab:abl_realisation}
\begin{adjustbox}{max width=\linewidth}
\renewcommand{\arraystretch}{1.15}
\begin{tabular}{@{}l|cc|cc@{}}
\toprule
\multicolumn{1}{@{}l!{\tableheaderupperseparator}}{\multirow{2}{*}{Perturbation}} & \multicolumn{2}{c}{OmniVinci} & \multicolumn{2}{c}{video-SALMONN~2+} \\
\cmidrule(lr){2-3}\cmidrule(lr){4-5}
\multicolumn{1}{@{}l!{\tableheaderlowerseparator}}{} & DAVE & \multicolumn{1}{c!{\tableheadersingleseparator}}{Daily-Omni} & DAVE & Daily-Omni \\
\midrule
Base & 44.3 & 61.2 & 42.5 & 61.6 \\
Pos. offset & 44.3\textsubscript{(+0.0)} & 61.2\textsubscript{\textcolor{mscdincrease}{(+0.1)}} & 43.7\textsubscript{\textcolor{mscdincrease}{(+1.2)}} & 62.7\textsubscript{\textcolor{mscdincrease}{(+1.2)}} \\
Random perm. & 47.6\textsubscript{\textcolor{mscdincrease}{(+3.3)}} & 62.1\textsubscript{\textcolor{mscdincrease}{(+0.9)}} & 42.7\textsubscript{\textcolor{mscdincrease}{(+0.2)}} & 63.2\textsubscript{\textcolor{mscdincrease}{(+1.7)}} \\
\rowcolor{mscdhighlight}
Cyclic shift & \textbf{49.8}\textsubscript{\textcolor{mscdincrease}{(+5.5)}} & \textbf{63.3}\textsubscript{\textcolor{mscdincrease}{(+2.2)}} & \textbf{45.5}\textsubscript{\textcolor{mscdincrease}{(+3.0)}} & \textbf{64.4}\textsubscript{\textcolor{mscdincrease}{(+2.8)}} \\
\bottomrule
\end{tabular}
\end{adjustbox}
\end{table}

\begin{table}[t]
\centering
\small
\setlength{\tabcolsep}{4pt}
\caption{Ablation study on window duration (accuracy, \%). 30-second and whole-clip windows outperform 10- or 20-second windows. Whole uses a single window spanning the entire clip. Subscripts show changes from Base in pp, and shading marks the default setting.}
\label{tab:abl_window_avut}
\begin{adjustbox}{max width=\linewidth}
\renewcommand{\arraystretch}{1.15}
\setlength{\tabcolsep}{3pt}
\begin{tabular}{@{}l|cccc@{}}
\toprule
\multicolumn{1}{@{}l!{\tableheaderupperseparator}}{\multirow{2}{*}{Window}} & \multicolumn{2}{c}{OmniVinci} & \multicolumn{2}{c}{video-SALMONN~2+} \\
\cmidrule(lr){2-3}\cmidrule(lr){4-5}
\multicolumn{1}{@{}l!{\tableheaderlowerseparator}}{} & DAVE & Daily-Omni & DAVE & Daily-Omni \\
\midrule
Base & 44.3 & 61.2 & 42.5 & 61.6 \\
10 s & 45.7\textsubscript{\textcolor{mscdincrease}{(+1.5)}} & 61.5\textsubscript{\textcolor{mscdincrease}{(+0.3)}} & 42.7\textsubscript{\textcolor{mscdincrease}{(+0.2)}} & 62.7\textsubscript{\textcolor{mscdincrease}{(+1.1)}} \\
20 s & 47.9\textsubscript{\textcolor{mscdincrease}{(+3.6)}} & 61.2\textsubscript{\textcolor{mscdincrease}{(+0.1)}} & 45.2\textsubscript{\textcolor{mscdincrease}{(+2.7)}} & 63.7\textsubscript{\textcolor{mscdincrease}{(+2.2)}} \\
\rowcolor{mscdhighlight}
30 s & 49.8\textsubscript{\textcolor{mscdincrease}{(+5.5)}} & \textbf{63.3}\textsubscript{\textcolor{mscdincrease}{(+2.2)}} & \textbf{45.5}\textsubscript{\textcolor{mscdincrease}{(+3.0)}} & 64.4\textsubscript{\textcolor{mscdincrease}{(+2.8)}} \\
Whole & \textbf{50.4}\textsubscript{\textcolor{mscdincrease}{(+6.1)}} & 62.7\textsubscript{\textcolor{mscdincrease}{(+1.5)}} & \textbf{45.5}\textsubscript{\textcolor{mscdincrease}{(+3.0)}} & \textbf{64.5}\textsubscript{\textcolor{mscdincrease}{(+2.9)}} \\
\bottomrule
\end{tabular}
\end{adjustbox}
\end{table}

MSCD aims to emphasize the contribution of audio--visual temporal alignment to the model's predictions.
However, an audio shift or a video shift not only disrupts audio--visual temporal alignment but also rearranges the content of the shifted modality.
We therefore model the logit differences between the original and single-shift conditions using the following additive approximation:
\begin{equation}
\begin{aligned}
    \mathbf z-\mathbf z_A &= \mathbf a_A+\mathbf o_A,\\
    \mathbf z-\mathbf z_V &= \mathbf a_V+\mathbf o_V,
\end{aligned}
\label{eq:additive_approximation}
\end{equation}
where $\mathbf a_A$ and $\mathbf a_V$ denote the contributions of audio--visual temporal alignment to these logit differences, and $\mathbf o_A$ and $\mathbf o_V$ denote the contributions of each modality's own temporal ordering.
To estimate the combined intra-modal order contribution, we additionally use an audio--video shift, which rearranges both modalities while preserving their temporal alignment.
Assuming that $\mathbf o_A$ and $\mathbf o_V$ are shared with the corresponding single shifts and combine additively, its logit difference is
\begin{equation}
    \mathbf z-\mathbf z_{AV}=\mathbf o_A+\mathbf o_V.
\label{eq:joint_order_contribution}
\end{equation}
Subtracting this difference from the sum of the two single-shift differences motivates the following contrastive term:
\begin{equation}
\begin{aligned}
    \Delta\mathbf z
    &= \tfrac12\bigl[(\mathbf z-\mathbf z_A)+(\mathbf z-\mathbf z_V)-(\mathbf z-\mathbf z_{AV})\bigr]\\
    &= \tfrac12\bigl(\mathbf z-\mathbf z_A-\mathbf z_V+\mathbf z_{AV}\bigr).
\end{aligned}
\label{eq:logit_correction}
\end{equation}
Under this additive approximation, $\Delta\mathbf z=(\mathbf a_A+\mathbf a_V)/2$, leaving only the cross-modal alignment contributions.
We add this contrastive term to the original logits to obtain
\begin{equation}
    \widetilde{\mathbf z}=\mathbf z+\alpha\Delta\mathbf z,
\label{eq:corrected_logits}
\end{equation}
where $\alpha\geq0$ controls the contrastive strength.

\section{Experiments}
\label{sec:experiments}

\subsection{Experimental setup}

\paragraph{Benchmarks.}
We evaluate our method on three benchmarks for audio--visual understanding using a common multiple-choice evaluation protocol.
For DAVE~\cite{NEURIPS2025_4d5f03fd}, we focus on \emph{Multimodal synchronization} (2,022 questions), which requires identifying the visual action occurring when a queried sound is heard.
Daily-Omni~\cite{zhou2026dailyomniaudiovisualreasoningtemporal} evaluates broader audio--visual understanding and reasoning in everyday scenarios.
We use all 1,197 questions across its six subtasks: AV Event Alignment, Comparative, Context Understanding, Event Sequence, Inference, and Reasoning.
For AVUT~\cite{yang-etal-2025-audio-centric}, we evaluate all three multiple-choice tasks in the human-annotated \emph{Audio Visual Alignment} subset, comprising 1,120 questions.
These tasks assess whether models can match spoken information to characters (AVCM), objects (AVOM), and on-screen text (AVTM).

\paragraph{Models.}
We evaluate MSCD on two audio--visual LLMs, OmniVinci and video-SALMONN~2+.
OmniVinci builds on Qwen2.5~\cite{qwen2025qwen25technicalreport}, grouping audio and visual embeddings by time and incorporating explicit time embeddings to represent their temporal relationships~\cite{ICLR2026_5c1863f7}.
video-SALMONN~2+ extends Qwen2.5-VL~\cite{bai2025qwen25vltechnicalreport} with an audio encoder and interleaves audio tokens with the corresponding visual tokens~\cite{tang2025video}.

\paragraph{Implementation details.}
\label{sec:implementation}
Our method uses a temporal window duration of $w=30$ seconds.
Shifts are applied before time embedding in OmniVinci and at the decoder input in video-SALMONN~2+.
We select one contrastive strength $\alpha$ per backbone by maximizing the unweighted mean accuracy across the three evaluation benchmarks.
The search covers $\alpha$ values from 0 to 6 in steps of 0.5.
This yields $\alpha=5.0$ for OmniVinci and $\alpha=1.5$ for video-SALMONN~2+.
The effect of $\alpha$ is analyzed in \cref{sec:ablations}.
For all methods and benchmarks, we select the answer choice with the highest probability at the first decoding step.

\paragraph{Baselines.}
We compare MSCD with four training-free inference methods for AV-LLMs: AVCD~\cite{NEURIPS2025_5b5e70e7}, MAD~\cite{Chung_2026_CVPR}, CAD~\cite{deng2026cad}, and ASD~\cite{pmlr-v306-jung26c}.
These baselines focus on mitigating cross-modal hallucinations, whereas MSCD aims to improve temporally grounded audio--visual understanding.
We adapt all baselines to both backbones and keep their hyperparameters fixed across benchmarks.
Implementation details are provided in \cref{sec:supp_baselines}.

\begin{figure*}[t]
    \centering
    \includegraphics[width=0.9\textwidth]{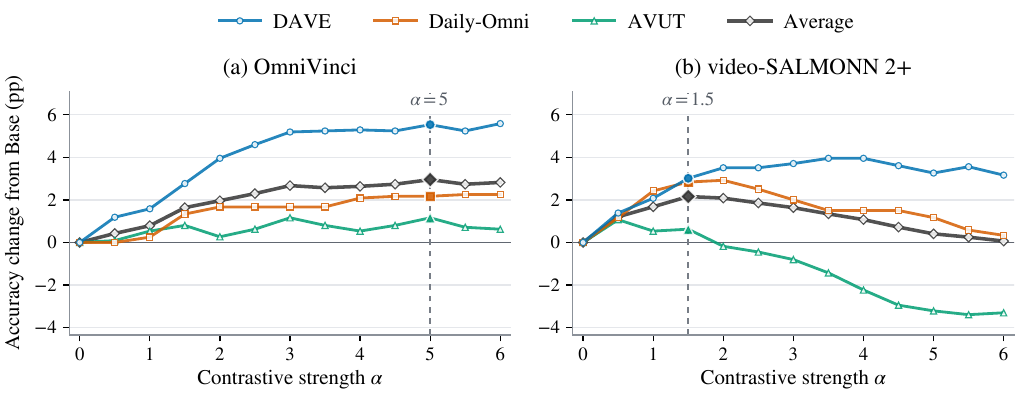}
    \caption{Effect of the contrastive strength $\alpha$. Curves show accuracy changes from Base (pp), and Average denotes the unweighted mean across benchmarks. Vertical dashed lines mark the strengths used in the main results.}
    \label{fig:alpha_sensitivity}
\end{figure*}

\subsection{Main results}
\label{sec:main_results}

\Cref{tab:main_w30p2_permodel} compares MSCD with Base and the four decoding baselines.
Among the evaluated methods, MSCD is the only one that improves accuracy over Base on all three benchmarks for both backbones.

MSCD improves performance on tasks that explicitly evaluate audio--visual temporal alignment.
On DAVE, OmniVinci reaches 49.8\% accuracy, exceeding Base by 5.5~pp, while video-SALMONN~2+ rises from 42.5\% to 45.5\%.
Both backbones also benefit from MSCD on Daily-Omni's AV Event Alignment subtask.
Together, these results support MSCD's effectiveness in identifying temporally corresponding audio and visual events.

These benefits extend beyond explicit temporal alignment to broader audio--visual understanding.
On Daily-Omni, MSCD achieves the highest overall accuracy for both backbones, exceeding Base by 2.2~pp with OmniVinci and 2.8~pp with video-SALMONN~2+.
None of the other decoding methods exceeds Base overall with either backbone.
MSCD matches or exceeds Base across all six subtasks, with both backbones gaining 4.1~pp on Context Understanding.

On AVUT, OmniVinci reaches 67.7\% overall accuracy, the highest among the compared methods.
For video-SALMONN~2+, MSCD raises accuracy from 61.8\% to 62.4\% but remains below MAD and CAD.
While character-matching accuracy remains unchanged, both backbones benefit on object and text matching.
These gains suggest that emphasizing temporal alignment can also help models establish semantic associations between audio and visual content.

\subsection{Ablations}
\label{sec:ablations}

\paragraph{Contrastive formulation.}
\Cref{tab:abl_contrast} compares MSCD with variants using the audio-shift contrast, the video-shift contrast, and their mean.
MSCD achieves the highest accuracy on all three benchmarks for both backbones.

The audio-shift contrast improves accuracy on DAVE for both backbones, but its benefits do not extend consistently to the other benchmarks.
The video-shift contrast falls below Base except on Daily-Omni with video-SALMONN~2+.
The \emph{Mean} variant averages the two single-shift contrasts without using the audio--video shift and falls below Base on all three benchmarks for OmniVinci.
Relative to \emph{Mean}, incorporating the audio--video shift raises OmniVinci's Daily-Omni accuracy from 58.1\% to 63.3\%.
These results support using the audio--video shift to mitigate the effects of rearrangement within each modality.

\paragraph{Temporal perturbation strategy.}
To assess the effect of perturbation design, we evaluate three strategies within MSCD (\cref{tab:abl_realisation}).
\emph{Positional offset} keeps the feature content and order unchanged and shifts only the temporal information assigned to a modality's features.
\emph{Random permutation} randomly reorders temporal blocks within each window.
\emph{Cyclic shift} swaps the two temporal halves within each window while preserving the order within each half.
Implementation details are given in \cref{sec:supp_ablation}.

Random permutation outperforms Base on both benchmarks with both backbones, while positional offset raises accuracy for video-SALMONN~2+ but produces little change for OmniVinci.
These results indicate that MSCD can benefit from different temporal perturbations.
Among the evaluated strategies, cyclic shift consistently yields the highest accuracy on DAVE and Daily-Omni for both backbones.
This advantage may stem from differences in how the perturbations alter temporal structure.
Random permutation can introduce frequent discontinuities in the reordered sequences, which may make the intra-modal order contributions harder to cancel out.
Positional offset can leave temporal positions near the sequence boundaries without features from one or both modalities, which may affect predictions beyond the intended change in audio--visual alignment.
Cyclic shift helps mitigate both issues.

\paragraph{Window duration.}
\label{sec:window_ablation}
We compare different window durations on DAVE and Daily-Omni using both backbones (\cref{tab:abl_window_avut}).
With a fixed window duration, the target temporal displacement is $w/2$ seconds for every full window, regardless of video length.
However, shorter windows introduce more frequent temporal discontinuities, which may degrade performance.
MSCD outperforms Base across all tested settings, and windows of 30 seconds yield higher accuracy than those of 10 or 20 seconds in all four settings.
Compared with windows of 30 seconds, whole-clip windows raise OmniVinci's accuracy on DAVE but lower it on Daily-Omni, while video-SALMONN~2+ performs similarly under both settings.

\paragraph{Contrastive strength.}
We examine how the contrastive strength affects accuracy across the three benchmarks (\cref{fig:alpha_sensitivity}).
Both backbones outperform Base on DAVE and Daily-Omni across a broad range of contrastive strengths.
However, the best-performing contrastive strength varies across benchmarks, with AVUT favoring smaller values, particularly for video-SALMONN~2+.

\begin{table}[t]
\centering
\small
\setlength{\tabcolsep}{4pt}
\caption{Incorrect-action selection rates (\%) on DAVE by temporal distance from the correct action. $N$ counts the incorrect options in each interval, and rates indicate the percentage selected by the model. Subscripts show changes from Base in pp.}
\label{tab:analysis_dave_temporal_errors}
\begin{adjustbox}{max width=\linewidth}
\renewcommand{\arraystretch}{1.15}
\setlength{\tabcolsep}{3pt}
\begin{tabular}{@{}lc|cc|cc@{}}
\toprule
\multirow{2}{*}{Distance (s)} & \multicolumn{1}{c!{\tableheaderupperseparator}}{\multirow{2}{*}{$N$}} & \multicolumn{2}{c!{\tableheaderupperseparator}}{OmniVinci} & \multicolumn{2}{c}{video-SALMONN~2+} \\
\cmidrule(lr){3-4}\cmidrule(lr){5-6}
 & \multicolumn{1}{c!{\tableheaderlowerseparator}}{} & Base & \multicolumn{1}{c!{\tableheaderlowerseparator}}{MSCD} & Base & MSCD \\
\midrule
$[0,4)$ & 357 & 11.8 & 13.7\textsubscript{(+2.0)} & 14.6 & 14.3\textsubscript{(-0.3)} \\
$[4,8)$ & 2,657 & 16.3 & 16.5\textsubscript{(+0.2)} & 16.6 & 16.6\textsubscript{(+0.1)} \\
$[8,12)$ & 1,429 & 20.2 & 16.5\textsubscript{(-3.7)} & 18.3 & 17.1\textsubscript{(-1.3)} \\
$[12,16)$ & 958 & 19.2 & 14.3\textsubscript{(-4.9)} & 18.5 & 15.6\textsubscript{(-2.9)} \\
$[16,20)$ & 397 & 23.9 & 15.6\textsubscript{(-8.3)} & 18.1 & 16.1\textsubscript{(-2.0)} \\
$\geq 20$ & 268 & 15.7 & 11.6\textsubscript{(-4.1)} & 17.2 & 14.9\textsubscript{(-2.2)} \\
\bottomrule
\end{tabular}
\end{adjustbox}
\end{table}

\begin{table}[t]
\centering
\small
\setlength{\tabcolsep}{4pt}
\caption{Effect of MSCD on the audio-delay probe (OmniVinci, DAVE). MSCD increases the selection rate of the shifted correct answer at all tested shift magnitudes. $N$ denotes the number of evaluated examples from both shift directions combined. Subscripts show changes from Base in pp.}
\label{tab:analysis_delay_mscd}
\begin{adjustbox}{max width=\linewidth}
\renewcommand{\arraystretch}{1.15}
\sbox{\tableheaderlabelbox}{\shortstack{Shifted correct\\answer (\%) $\uparrow$}\shortstack{Original correct\\answer (\%)}}
\begin{tabular}{@{}cc|cc|cc@{}}
\toprule
\multirow{2}{*}{Delay (s)} & \multicolumn{1}{c!{\tableheaderupperseparator[\ht\tableheaderlabelbox]}}{\multirow{2}{*}{$N$}} & \multicolumn{2}{c!{\tableheaderupperseparator[\ht\tableheaderlabelbox]}}{\shortstack{Shifted correct\\answer (\%) $\uparrow$}} & \multicolumn{2}{c}{\shortstack{Original correct\\answer (\%)}} \\
\cmidrule(lr){3-4}\cmidrule(lr){5-6}
 & \multicolumn{1}{c!{\tableheaderlowerseparator}}{} & Base & \multicolumn{1}{c!{\tableheaderlowerseparator}}{MSCD} & Base & MSCD \\
\midrule
$\pm4$ & 603 & 22.7 & \textbf{25.7}\textsubscript{\textcolor{mscdincrease}{(+3.0)}} & 35.7 & 37.0\textsubscript{(+1.3)} \\
$\pm8$ & 1457 & 25.0 & \textbf{29.6}\textsubscript{\textcolor{mscdincrease}{(+4.6)}} & 38.0 & 37.7\textsubscript{(-0.2)} \\
$\pm16$ & 938 & 29.4 & \textbf{35.6}\textsubscript{\textcolor{mscdincrease}{(+6.2)}} & 34.3 & 26.9\textsubscript{(-7.5)} \\
\bottomrule
\end{tabular}
\end{adjustbox}
\end{table}

\providecommand{\trackingobservationref}{\cref{sec:obs_delay}}
\providecommand{\trackingdirectionref}{\cref{sec:supp_delay_results}}
\providecommand{\temporaltrackingpreviewbreak}{}

\subsection{Temporal behavior of MSCD}
\label{sec:temporal_analysis_draft}

To examine how MSCD changes the model's errors, we analyze incorrect-action selections on the original DAVE inputs (\cref{tab:analysis_dave_temporal_errors}).
We collect the three incorrect action options from each question and group them by their temporal distance from the correct action.
We define this distance as the absolute difference between the midpoints of the correct and incorrect action intervals.
For each distance interval, we report the percentage of these options selected by the model.
In this setting, MSCD reduces the selection rate of incorrect actions at least 8 seconds away from the correct action in every reported interval.
For example, in the 12--16-second interval, the selection rate decreases from 19.2\% to 14.3\% for OmniVinci and from 18.5\% to 15.6\% for video-SALMONN~2+.
Changes below 8 seconds are small or mixed.
With MSCD, selected incorrect actions lie closer to the correct action on average: the mean temporal distance decreases from 9.98 to 9.27 seconds for OmniVinci and from 9.77 to 9.50 seconds for video-SALMONN~2+.

\temporaltrackingpreviewbreak
To test whether MSCD also follows changes in audio--visual timing, we revisit the audio-delay probe in \trackingobservationref{} using OmniVinci.
With examples from both shift directions combined, MSCD increases the selection rate of the \emph{shifted correct answer} at all three tested magnitudes (\cref{tab:analysis_delay_mscd}).
For a displacement of 16 seconds, the shifted correct answer is selected in 35.6\% of cases, compared with 26.9\% for the \emph{original correct answer}, reversing the ordering observed for Base.
Together, these analyses suggest that MSCD helps models use audio--visual timing to select answers, with reduced confusion between temporally distant actions.

\section{Conclusion}
\label{sec:conclusion}

In this work, we present Modality-Shift Contrastive Decoding (MSCD), a training-free decoding method for temporally grounded audio--visual understanding.
MSCD contrasts logits under preserved and disrupted temporal alignment, using the audio--video shift to mitigate the effects of rearranging each modality's content.
In experiments with two AV-LLM backbones on three benchmarks, MSCD consistently enhances accuracy over the base models, supporting temporal alignment as a useful cue for joint audio--visual understanding.
The audio-shift probe further provides behavioral evidence that MSCD can better follow changes in audio--visual temporal alignment.

\section*{LLM Usage Statement}

We used Claude 5.1 Fable primarily to assist with code generation and code review for the experimental implementation, and GPT-6 Astra primarily to assist with drafting, editing, and refining the manuscript.
LLM assistance also supported literature searches, the preparation of figures and tables, and discussions of experimental ideas and the interpretation of results.
The authors made the final research decisions and take full responsibility for the implementation, experimental results, and manuscript.

{
    \small
    \bibliographystyle{ieeenat_fullname}
    \bibliography{main}

\begin{thebibliography}{48}
\providecommand{\natexlab}[1]{#1}
\providecommand{\url}[1]{\texttt{#1}}
\expandafter\ifx\csname urlstyle\endcsname\relax
  \providecommand{\doi}[1]{doi: #1}\else
  \providecommand{\doi}{doi: \begingroup \urlstyle{rm}\Url}\fi

\bibitem[Bai et~al.(2025)Bai, Chen, Liu, Wang, Ge, Song, Dang, Wang, Wang,
  Tang, Zhong, Zhu, Yang, Li, Wan, Wang, Ding, Fu, Xu, Ye, Zhang, Xie, Cheng,
  Zhang, Yang, Xu, and Lin]{bai2025qwen25vltechnicalreport}
Shuai Bai, Keqin Chen, Xuejing Liu, Jialin Wang, Wenbin Ge, Sibo Song, Kai
  Dang, Peng Wang, Shijie Wang, Jun Tang, Humen Zhong, Yuanzhi Zhu, Mingkun
  Yang, Zhaohai Li, Jianqiang Wan, Pengfei Wang, Wei Ding, Zheren Fu, Yiheng
  Xu, Jiabo Ye, Xi Zhang, Tianbao Xie, Zesen Cheng, Hang Zhang, Zhibo Yang,
  Haiyang Xu, and Junyang Lin.
\newblock Qwen2.5-vl technical report, 2025.

\bibitem[Cheng et~al.(2024)Cheng, Leng, Zhang, Xin, Li, Chen, Zhu, Zhang, Luo,
  Zhao, et~al.]{cheng2024videollama}
Zesen Cheng, Sicong Leng, Hang Zhang, Yifei Xin, Xin Li, Guanzheng Chen,
  Yongxin Zhu, Wenqi Zhang, Ziyang Luo, Deli Zhao, et~al.
\newblock Videollama 2: Advancing spatial-temporal modeling and audio
  understanding in video-llms.
\newblock \emph{arXiv preprint arXiv:2406.07476}, 2024.

\bibitem[Chowdhury et~al.(2024)Chowdhury, Nag, Dasgupta, Chen, Elhoseiny, Gao,
  and Manocha]{chowdhury2024meerkat}
Sanjoy Chowdhury, Sayan Nag, Subhrajyoti Dasgupta, Jun Chen, Mohamed Elhoseiny,
  Ruohan Gao, and Dinesh Manocha.
\newblock Meerkat: Audio-visual large language model for grounding in space and
  time.
\newblock In \emph{European Conference on Computer Vision}, pages 52--70.
  Springer, 2024.

\bibitem[Chung et~al.(2026)Chung, Kim, Chee, and Ro]{Chung_2026_CVPR}
Sangyun Chung, Se~Yeon Kim, Youngchae Chee, and Yong~Man Ro.
\newblock Mad: Modality-adaptive decoding for mitigating cross-modal
  hallucinations in multimodal large language models.
\newblock In \emph{Proceedings of the IEEE/CVF Conference on Computer Vision
  and Pattern Recognition (CVPR)}, pages 40175--40185, 2026.

\bibitem[Dai et~al.(2026)Dai, Wu, Zeng, Hua, Liu, Li, Wang, Tong, Liang, Ma,
  Niu, Guo, Shi, Ding, Ji, Mei, Guan, Zhang, Wan, Fu, and
  Zhang]{dai2026latentomni}
Yifan Dai, Zhenhua Wu, Bohan Zeng, Daili Hua, Jialing Liu, Bozhou Li, Yuran
  Wang, Chengzhuo Tong, Hao Liang, Xiaochen Ma, Junbo Niu, Tianyu Guo, Yang
  Shi, Yue Ding, Yiyan Ji, Bingyin Mei, Yushuo Guan, Yuanxing Zhang, Pengfei
  Wan, Fangcheng Fu, and Wentao Zhang.
\newblock Latentomni: Rethinking omni-modal understanding via unified
  audio-visual latent reasoning.
\newblock In \emph{Advances in Neural Information Processing Systems}, 2026.

\bibitem[Deng et~al.(2026)Deng, Sun, Zheng, Yang, and Han]{deng2026cad}
Yuchen Deng, Chang Sun, Hai-Tao Zheng, Feidiao Yang, and Yuxing Han.
\newblock Cad: Conflict-aware decoding to mitigate cross-modal hallucinations
  in omnimodal large language models.
\newblock \emph{arXiv preprint arXiv:2609.04247}, 2026.

\bibitem[Fu et~al.(2024)Fu, Lin, Long, Shen, Dai, Zhao, Zhang, Dong, Li, Wang,
  et~al.]{fu2024vita}
Chaoyou Fu, Haojia Lin, Zuwei Long, Yunhang Shen, Yuhang Dai, Meng Zhao, Yi-Fan
  Zhang, Shaoqi Dong, Yangze Li, Xiong Wang, et~al.
\newblock Vita: Towards open-source interactive omni multimodal llm.
\newblock \emph{arXiv preprint arXiv:2408.05211}, 2024.

\bibitem[Ghosh et~al.(2026)Ghosh, Goel, Jayakumar, Koroshinadze, Anand,
  Gururani, Ye, Biswas, Su, Hosseini-Asl, et~al.]{ghosh2026audio}
Sreyan Ghosh, Arushi Goel, Kaousheik Jayakumar, Lasha Koroshinadze, Nishit
  Anand, Siddharth Gururani, Hanrong Ye, Pritam Biswas, Yuanhang Su, Ehsan
  Hosseini-Asl, et~al.
\newblock Audio-visual flamingo: Open audio-visual intelligence for long and
  complex videos.
\newblock \emph{arXiv preprint arXiv:2607.16107}, 2026.

\bibitem[Gong et~al.(2026)Gong, Feng, Li, Wang, Cheng, Yang, Han, Wang, Bai,
  Yang, and Yue]{gong-etal-2026-probing}
Kaixiong Gong, Kaituo Feng, Bohao Li, Yibing Wang, Mofan Cheng, Shijia Yang,
  Jiaming Han, Benyou Wang, Yutong Bai, Zhuoran Yang, and Xiangyu Yue.
\newblock Probing audio-visual reasoning in multimodal language models through
  the lens of audio.
\newblock In \emph{Proceedings of the 64th Annual Meeting of the {A}ssociation
  for {C}omputational {L}inguistics (Volume 1: Long Papers)}, pages
  36603--36645, San Diego, California, United States, 2026. Association for
  Computational Linguistics.

\bibitem[Han et~al.(2024)Han, Gong, Zhang, Wang, Zhang, Lin, Qiao, Gao, and
  Yue]{Han_2024_CVPR}
Jiaming Han, Kaixiong Gong, Yiyuan Zhang, Jiaqi Wang, Kaipeng Zhang, Dahua Lin,
  Yu Qiao, Peng Gao, and Xiangyu Yue.
\newblock Onellm: One framework to align all modalities with language.
\newblock In \emph{Proceedings of the IEEE/CVF Conference on Computer Vision
  and Pattern Recognition (CVPR)}, pages 26584--26595, 2024.

\bibitem[Hong et~al.(2026)Hong, Yan, Cai, Jiang, Hu, and
  Xie]{ICLR2026_55dc32df}
Jack Hong, Shilin Yan, Jiayin Cai, Xiaolong Jiang, Yao Hu, and Weidi Xie.
\newblock Worldsense: Evaluating real-world omnimodal understanding for
  multimodal llms.
\newblock In \emph{International Conference on Learning Representations}, pages
  52423--52443, 2026.

\bibitem[Jung et~al.(2025)Jung, Jang, and Chung]{NEURIPS2025_5b5e70e7}
Chaeyoung Jung, Youngjoon Jang, and Joon~Son Chung.
\newblock Avcd: Mitigating hallucinations in audio-visual large language models
  through contrastive decoding.
\newblock In \emph{Advances in Neural Information Processing Systems}, pages
  63143--63174. Curran Associates, Inc., 2025.

\bibitem[Jung et~al.(2026{\natexlab{a}})Jung, Jang, Choi, and
  Chung]{jung-etal-2026-fork}
Chaeyoung Jung, Youngjoon Jang, Jongmin Choi, and Joon~Son Chung.
\newblock Fork-merge decoding: Enhancing multimodal understanding in
  audio-visual large language models.
\newblock In \emph{Findings of the {A}ssociation for {C}omputational
  {L}inguistics: {EMNLP} 2026}, pages 29610--29629, Budapest, Hungary,
  2026{\natexlab{a}}. Association for Computational Linguistics.

\bibitem[Jung et~al.(2026{\natexlab{b}})Jung, Jung, Kim, and
  Chung]{pmlr-v306-jung26c}
Jihoo Jung, Chaeyoung Jung, Ji-Hoon Kim, and Joon~Son Chung.
\newblock Probing cross-modal information hubs in audio-visual {LLM}s.
\newblock In \emph{Proceedings of the 43rd International Conference on Machine
  Learning}, pages 54928--54952. PMLR, 2026{\natexlab{b}}.

\bibitem[Korbar et~al.(2018)Korbar, Tran, and Torresani]{NEURIPS2018_c4616f5a}
Bruno Korbar, Du Tran, and Lorenzo Torresani.
\newblock Cooperative learning of audio and video models from self-supervised
  synchronization.
\newblock In \emph{Advances in Neural Information Processing Systems}. Curran
  Associates, Inc., 2018.

\bibitem[Leng et~al.(2024)Leng, Zhang, Chen, Li, Lu, Miao, and
  Bing]{Leng_2024_CVPR}
Sicong Leng, Hang Zhang, Guanzheng Chen, Xin Li, Shijian Lu, Chunyan Miao, and
  Lidong Bing.
\newblock Mitigating object hallucinations in large vision-language models
  through visual contrastive decoding.
\newblock In \emph{Proceedings of the IEEE/CVF Conference on Computer Vision
  and Pattern Recognition (CVPR)}, pages 13872--13882, 2024.

\bibitem[Li et~al.(2022)Li, Wei, Tian, Xu, Wen, and Hu]{Li_2022_CVPR}
Guangyao Li, Yake Wei, Yapeng Tian, Chenliang Xu, Ji-Rong Wen, and Di Hu.
\newblock Learning to answer questions in dynamic audio-visual scenarios.
\newblock In \emph{Proceedings of the IEEE/CVF Conference on Computer Vision
  and Pattern Recognition (CVPR)}, pages 19108--19118, 2022.

\bibitem[Li et~al.(2023)Li, Holtzman, Fried, Liang, Eisner, Hashimoto,
  Zettlemoyer, and Lewis]{li-etal-2023-contrastive}
Xiang~Lisa Li, Ari Holtzman, Daniel Fried, Percy Liang, Jason Eisner, Tatsunori
  Hashimoto, Luke Zettlemoyer, and Mike Lewis.
\newblock Contrastive decoding: Open-ended text generation as optimization.
\newblock In \emph{Proceedings of the 61st Annual Meeting of the {A}ssociation
  for {C}omputational {L}inguistics (Volume 1: Long Papers)}, pages
  12286--12312, Toronto, Canada, 2023. Association for Computational
  Linguistics.

\bibitem[Li et~al.(2025{\natexlab{a}})Li, Liu, Zhang, Chen, Li, Li, Liu, Ming,
  Dong, Pan, et~al.]{li2025baichuan}
Yadong Li, Jun Liu, Tao Zhang, Song Chen, Tianpeng Li, Zehuan Li, Lijun Liu,
  Lingfeng Ming, Guosheng Dong, Da Pan, et~al.
\newblock Baichuan-omni-1.5 technical report.
\newblock \emph{arXiv preprint arXiv:2501.15368}, 2025{\natexlab{a}}.

\bibitem[Li et~al.(2025{\natexlab{b}})Li, Zhang, Ma, Yuan, Zhu, Guo, Liang,
  Liu, Wang, Yang, Wu, Qu, Shi, Zhang, Yang, WEN, Wang, Li, ZHANG, Liu,
  Benetos, Huang, and Lin]{NEURIPS2025_2c7c4a12}
Yizhi Li, Ge Zhang, Yinghao Ma, Ruibin Yuan, Zhu, Hangyu Guo, Yiming Liang,
  Jiaheng Liu, Noah Wang, Jian Yang, Siwei Wu, Xingwei Qu, Jinjie Shi, Xinyue
  Zhang, Zhenzhu Yang, Yidan WEN, Yanghai Wang, Shihao Li, ZHAO-XIANG ZHANG,
  Ruibo Liu, Emmanouil Benetos, Wenhao Huang, and Chenghua Lin.
\newblock Omnibench: Towards the future of universal omni-language models.
\newblock In \emph{Advances in Neural Information Processing Systems}. Curran
  Associates, Inc., 2025{\natexlab{b}}.

\bibitem[Li et~al.(2026)Li, Liu, Song, Wei, Tak{\'a}{\v{c}}, and
  Lahlou]{li-etal-2026-temporal}
Yanda Li, Yuhan Liu, Zirui Song, Yunchao Wei, Martin Tak{\'a}{\v{c}}, and Salem
  Lahlou.
\newblock Temporal contrastive decoding: A training-free method for large
  audio-language models.
\newblock In \emph{Findings of the {A}ssociation for {C}omputational
  {L}inguistics: {ACL} 2026}, pages 14921--14934, San Diego, California, United
  States, 2026. Association for Computational Linguistics.

\bibitem[Liu et~al.(2025)Liu, Dong, Wang, Liu, Hu, Lu, and Rao]{liu2025ola}
Zuyan Liu, Yuhao Dong, Jiahui Wang, Ziwei Liu, Winston Hu, Jiwen Lu, and
  Yongming Rao.
\newblock Ola: Pushing the frontiers of omni-modal language model.
\newblock \emph{arXiv preprint arXiv:2502.04328}, 2025.

\bibitem[Lu et~al.(2026)Lu, Yu, Li, Zhang, Ye, and Chang]{Lu_2026_CVPR}
Weiheng Lu, An Yu, Jian Li, Zhenfei Zhang, Felix X.-F. Ye, and Ming-Ching
  Chang.
\newblock Fave: A structured benchmark for fine-grained audio-visual temporal
  evaluation in multimodal llms.
\newblock In \emph{Proceedings of the IEEE/CVF Conference on Computer Vision
  and Pattern Recognition (CVPR)}, pages 1651--1660, 2026.

\bibitem[Panagopoulou et~al.(2024)Panagopoulou, Xue, Yu, Li, Li, Joty, Xu,
  Savarese, Xiong, and Niebles]{panagopoulou2024x}
Artemis Panagopoulou, Le Xue, Ning Yu, Junnan Li, Dongxu Li, Shafiq Joty, Ran
  Xu, Silvio Savarese, Caiming Xiong, and Juan~Carlos Niebles.
\newblock X-instructblip: A framework for aligning image, 3d, audio, video to
  llms and its emergent cross-modal reasoning.
\newblock In \emph{European Conference on Computer Vision}, pages 177--197.
  Springer, 2024.

\bibitem[Qwen et~al.(2025)Qwen, Yang, Yang, Zhang, Hui, Zheng, Yu, Li, Liu,
  Huang, Wei, Lin, Yang, Tu, Zhang, Yang, Yang, Zhou, Lin, Dang, Lu, Bao, Yang,
  Yu, Li, Xue, Zhang, Zhu, Men, Lin, Li, Tang, Xia, Ren, Ren, Fan, Su, Zhang,
  Wan, Liu, Cui, Zhang, and Qiu]{qwen2025qwen25technicalreport}
Qwen, An Yang, Baosong Yang, Beichen Zhang, Binyuan Hui, Bo Zheng, Bowen Yu,
  Chengyuan Li, Dayiheng Liu, Fei Huang, Haoran Wei, Huan Lin, Jian Yang,
  Jianhong Tu, Jianwei Zhang, Jianxin Yang, Jiaxi Yang, Jingren Zhou, Junyang
  Lin, Kai Dang, Keming Lu, Keqin Bao, Kexin Yang, Le Yu, Mei Li, Mingfeng Xue,
  Pei Zhang, Qin Zhu, Rui Men, Runji Lin, Tianhao Li, Tianyi Tang, Tingyu Xia,
  Xingzhang Ren, Xuancheng Ren, Yang Fan, Yang Su, Yichang Zhang, Yu Wan,
  Yuqiong Liu, Zeyu Cui, Zhenru Zhang, and Zihan Qiu.
\newblock Qwen2.5 technical report, 2025.

\bibitem[Ra et~al.(2026)Ra, Jung, Park, and Hong]{ra2026relevant}
Hyunjae Ra, Aecheon Jung, Jungin Park, and Sungeun Hong.
\newblock Relevant evidence decoding for audio-visual hallucination mitigation.
\newblock \emph{arXiv preprint arXiv:2610.02976}, 2026.

\bibitem[Radevski et~al.(2025)Radevski, Popordanoska, Blaschko, and
  Tuytelaars]{NEURIPS2025_4d5f03fd}
Gorjan Radevski, Teodora Popordanoska, Matthew Blaschko, and Tinne Tuytelaars.
\newblock Dave: Diagnostic benchmark for audio visual evaluation.
\newblock In \emph{Advances in Neural Information Processing Systems}. Curran
  Associates, Inc., 2025.

\bibitem[Radford et~al.(2023)Radford, Kim, Xu, Brockman, Mcleavey, and
  Sutskever]{pmlr-v202-radford23a}
Alec Radford, Jong~Wook Kim, Tao Xu, Greg Brockman, Christine Mcleavey, and
  Ilya Sutskever.
\newblock Robust speech recognition via large-scale weak supervision.
\newblock In \emph{Proceedings of the 40th International Conference on Machine
  Learning}, pages 28492--28518. PMLR, 2023.

\bibitem[Su et~al.(2024)Su, Ahmed, Lu, Pan, Bo, and Liu]{SU2024127063}
Jianlin Su, Murtadha Ahmed, Yu Lu, Shengfeng Pan, Wen Bo, and Yunfeng Liu.
\newblock Roformer: Enhanced transformer with rotary position embedding.
\newblock \emph{Neurocomputing}, 568:\penalty0 127063, 2024.

\bibitem[Su et~al.(2023)Su, Lan, Li, Xu, Wang, and Cai]{su-etal-2023-pandagpt}
Yixuan Su, Tian Lan, Huayang Li, Jialu Xu, Yan Wang, and Deng Cai.
\newblock {P}anda{GPT}: One model to instruction-follow them all.
\newblock In \emph{Proceedings of the 1st Workshop on Taming Large Language
  Models: Controllability in the era of Interactive Assistants}, pages 11--23,
  Prague, Czech Republic, 2023. Association for Computational Linguistics.

\bibitem[Sun et~al.(2024)Sun, Yu, Tang, Chen, Tan, Li, Lu, Ma, Wang, and
  Zhang]{pmlr-v235-sun24l}
Guangzhi Sun, Wenyi Yu, Changli Tang, Xianzhao Chen, Tian Tan, Wei Li, Lu Lu,
  Zejun Ma, Yuxuan Wang, and Chao Zhang.
\newblock video-{SALMONN}: Speech-enhanced audio-visual large language models.
\newblock In \emph{Proceedings of the 41st International Conference on Machine
  Learning}, pages 47198--47217. PMLR, 2024.

\bibitem[Tang et~al.(2025)Tang, Li, Yang, Zhuang, Sun, Li, Ma, and
  Zhang]{tang2025video}
Changli Tang, Yixuan Li, Yudong Yang, Jimin Zhuang, Guangzhi Sun, Wei Li, Zejun
  Ma, and Chao Zhang.
\newblock video-salmonn 2: Caption-enhanced audio-visual large language models.
\newblock \emph{arXiv preprint arXiv:2506.15220}, 2025.

\bibitem[Team(2026)]{team2026qwen3}
Qwen Team.
\newblock Qwen3. 5-omni technical report.
\newblock \emph{arXiv preprint arXiv:2604.15804}, 2026.

\bibitem[Tian et~al.(2018)Tian, Shi, Li, Duan, and Xu]{tian2018audio}
Yapeng Tian, Jing Shi, Bochen Li, Zhiyao Duan, and Chenliang Xu.
\newblock Audio-visual event localization in unconstrained videos.
\newblock In \emph{European conference on computer vision}, pages 252--268.
  Springer, 2018.

\bibitem[Wen et~al.(2026)Wen, Mo, Fu, Cai, Zhu, Li, Xie, Chen, and
  Qi]{wen2026whenvisionspeaksforsound}
Xiaofei Wen, Wenjie~Jacky Mo, Xingyu Fu, Rui Cai, Tinghui Zhu, Wendi Li, Yanan
  Xie, Muhao Chen, and Peng Qi.
\newblock When vision speaks for sound.
\newblock 2026.

\bibitem[Wu et~al.(2026)Wu, Chang, Sheng, Chung, Wang, and Wang]{Wu_2026_CVPR}
Chang-Hsun Wu, Kai-Po Chang, Yu-Yang Sheng, Hung-Kai Chung, Kuei-Chun Wang, and
  Yu-Chiang~Frank Wang.
\newblock Season: Mitigating temporal hallucination in video large language
  models via self-diagnostic contrastive decoding.
\newblock In \emph{Proceedings of the IEEE/CVF Conference on Computer Vision
  and Pattern Recognition (CVPR)}, pages 11096--11105, 2026.

\bibitem[Xu et~al.(2025{\natexlab{a}})Xu, Guo, He, Hu, He, Bai, Chen, Wang,
  Fan, Dang, Zhang, Wang, Chu, and Lin]{Qwen2.5-Omni}
Jin Xu, Zhifang Guo, Jinzheng He, Hangrui Hu, Ting He, Shuai Bai, Keqin Chen,
  Jialin Wang, Yang Fan, Kai Dang, Bin Zhang, Xiong Wang, Yunfei Chu, and
  Junyang Lin.
\newblock Qwen2.5-omni technical report.
\newblock \emph{arXiv preprint arXiv:2503.20215}, 2025{\natexlab{a}}.

\bibitem[Xu et~al.(2025{\natexlab{b}})Xu, Guo, Hu, Chu, Wang, He, Wang, Shi,
  He, Zhu, et~al.]{xu2025qwen3}
Jin Xu, Zhifang Guo, Hangrui Hu, Yunfei Chu, Xiong Wang, Jinzheng He, Yuxuan
  Wang, Xian Shi, Ting He, Xinfa Zhu, et~al.
\newblock Qwen3-omni technical report.
\newblock \emph{arXiv preprint arXiv:2509.17765}, 2025{\natexlab{b}}.

\bibitem[Xu et~al.(2026)Xu, Li, Hu, and Hu]{xu2026syncra}
Zelong Xu, Yan Li, Wenhe Hu, and Xiyang Hu.
\newblock Syncra: Learning temporal correspondence in omni-modal models.
\newblock \emph{arXiv preprint arXiv:2609.34363}, 2026.

\bibitem[Yang et~al.(2022)Yang, Wang, Duan, Chen, Hou, Jin, and
  Zhu]{10.1145/3503161.3548291}
Pinci Yang, Xin Wang, Xuguang Duan, Hong Chen, Runze Hou, Cong Jin, and Wenwu
  Zhu.
\newblock Avqa: A dataset for audio-visual question answering on videos.
\newblock In \emph{Proceedings of the 30th ACM International Conference on
  Multimedia}, page 3480–3491, New York, NY, USA, 2022. Association for
  Computing Machinery.

\bibitem[Yang et~al.(2025)Yang, Zhuang, Sun, Tang, Li, Li, Jiang, Li, Ma, and
  Zhang]{yang-etal-2025-audio-centric}
Yudong Yang, Jimin Zhuang, Guangzhi Sun, Changli Tang, Yixuan Li, Peihan Li,
  Yifan Jiang, Wei Li, Zejun Ma, and Chao Zhang.
\newblock Audio-centric video understanding benchmark without text shortcut.
\newblock In \emph{Proceedings of the 2025 Conference on Empirical Methods in
  Natural Language Processing}, pages 6569--6587, Suzhou, China, 2025.
  Association for Computational Linguistics.

\bibitem[Ye et~al.(2026)Ye, Yang, Goel, Huang, Wan, Tian, Cheng, Zhu, Su, Lou,
  Lin, Yang, Ghosh, Liu, Chen, Jahangiri, Dantrey, Xu, Hosseini-Asl,
  Mohseni~Taheri, Nariyambut~Murali, Liu, Lu, Olabiyi, Wang, Valle, Catanzaro,
  Tao, Han, Kautz, Yin, and Molchanov]{ICLR2026_5c1863f7}
Hanrong Ye, Chao-Han~Huck Yang, Arushi Goel, Wei Huang, Zhen Wan, Jinchuan
  Tian, An-Chieh Cheng, Ligeng Zhu, Yuanhang Su, Yuming Lou, Yong-Xiang Lin,
  Dong Yang, Sreyan Ghosh, Zhijian Liu, Yukang Chen, Ehsan Jahangiri, Ambrish
  Dantrey, Daguang Xu, Ehsan Hosseini-Asl, Seyed~Danial Mohseni~Taheri, Vidya
  Nariyambut~Murali, Sifei Liu, Yao Lu, Oluwatobi Olabiyi, Yu-Chiang~Frank
  Wang, Rafael Valle, Bryan Catanzaro, Andrew Tao, Song Han, Jan Kautz,
  Hongxu~(Danny) Yin, and Pavlo Molchanov.
\newblock Omnivinci: Enhancing architecture and data for omni-modal
  understanding llm.
\newblock In \emph{International Conference on Learning Representations}, pages
  56101--56138, 2026.

\bibitem[Ye et~al.(2024)Ye, Yu, Shao, Xie, Torr, and Cao]{ye2024cat}
Qilang Ye, Zitong Yu, Rui Shao, Xinyu Xie, Philip Torr, and Xiaochun Cao.
\newblock Cat: Enhancing multimodal large language model to answer questions in
  dynamic audio-visual scenarios.
\newblock In \emph{European Conference on Computer Vision}, pages 146--164.
  Springer, 2024.

\bibitem[Zhai et~al.(2023)Zhai, Mustafa, Kolesnikov, and Beyer]{Zhai_2023_ICCV}
Xiaohua Zhai, Basil Mustafa, Alexander Kolesnikov, and Lucas Beyer.
\newblock Sigmoid loss for language image pre-training.
\newblock In \emph{Proceedings of the IEEE/CVF International Conference on
  Computer Vision (ICCV)}, pages 11975--11986, 2023.

\bibitem[Zhang et~al.(2026)Zhang, Kim, Liu, and Oh]{Zhang_2026_SVHalluc}
Chenshuang Zhang, Kyeong~Seon Kim, Chengxin Liu, and Tae-Hyun Oh.
\newblock Svhalluc: Benchmarking speech--vision hallucination in audio-visual
  large language models.
\newblock In \emph{Proceedings of the IEEE/CVF Conference on Computer Vision
  and Pattern Recognition (CVPR)}, 2026.

\bibitem[Zhang et~al.(2023)Zhang, Li, and Bing]{zhang-etal-2023-video}
Hang Zhang, Xin Li, and Lidong Bing.
\newblock Video-{LL}a{MA}: An instruction-tuned audio-visual language model for
  video understanding.
\newblock In \emph{Proceedings of the 2023 Conference on Empirical Methods in
  Natural Language Processing: System Demonstrations}, pages 543--553,
  Singapore, 2023. Association for Computational Linguistics.

\bibitem[Zhao et~al.(2026)Zhao, Ma, Huang, Fan, Li, Kang, Wei, Yang, and
  Tai]{10.1007/978-3-032-37490-5_4}
Chen Zhao, Jiajun Ma, Qilong Huang, Tiehan Fan, Hongyu Li, Zhuoliang Kang,
  Xiaoming Wei, Jian Yang, and Ying Tai.
\newblock Temporal and cross-modal alignment for enhanced audiovisual video
  captioning.
\newblock In \emph{Computer Vision -- ECCV 2026}, pages 65--85, Cham, 2026.
  Springer Nature Switzerland.

\bibitem[Zhou et~al.(2026)Zhou, Wang, Wu, and
  Jiang]{zhou2026dailyomniaudiovisualreasoningtemporal}
Ziwei Zhou, Rui Wang, Zuxuan Wu, and Yu-Gang Jiang.
\newblock Daily-omni: Towards audio-visual reasoning with temporal alignment
  across modalities, 2026.

\end{thebibliography}
}
\clearpage
\setcounter{topnumber}{2}
\appendix
\maketitlesupplementary
\raggedbottom

\setcounter{table}{0}
\setcounter{figure}{0}
\setcounter{equation}{0}
\renewcommand{\thetable}{S\arabic{table}}
\renewcommand{\thefigure}{S\arabic{figure}}
\renewcommand{\theequation}{S\arabic{equation}}
\renewcommand{\theHtable}{supp.\arabic{table}}
\renewcommand{\theHfigure}{supp.\arabic{figure}}
\renewcommand{\theHequation}{supp.\arabic{equation}}

\newcommand{\supplementarytablehere}[1]{%
  \par\addvspace{\textfloatsep}\noindent
  \begin{minipage}{\columnwidth}
    \renewenvironment{table}[1][]{}{\par}
    \captionsetup{type=table,hypcap=false}
    \input{#1}
  \end{minipage}\par\addvspace{\textfloatsep}
}

\section{Implementation Details}
\label{sec:supp}

\subsection{Video sampling and spatial resolution}
\label{sec:supp_preprocessing}

For each backbone, Base and all decoding methods use the same video preprocessing across benchmarks, ablations, and the audio-delay probe.

\paragraph{OmniVinci.}
We uniformly sample 64 frames across each video and resize each frame to $448\times448$ pixels using the backbone's native SigLIP image processor~\cite{Zhai_2023_ICCV}.
With a fixed frame count, longer videos are sampled more sparsely in time.

\paragraph{video-SALMONN~2+.}
We sample frames at approximately one-second intervals.
If this yields more than 120 frames, we uniformly sample 120 frames across the full video.
Each frame is resized with a maximum pixel budget of 61,250, regardless of the number of sampled frames.

\subsection{Temporal shift implementation}
\label{sec:supp_shifts}

Temporal shifts are applied after encoding the audio and video, at different stages in the two backbones.
The resulting encoder outputs are reused across the four conditions, while each condition is processed separately by the decoder.
If the final window is shorter than $w$ seconds, we swap its two temporal halves.

\paragraph{OmniVinci.}
OmniVinci encodes timestamps in audio and visual features through constrained rotary time embedding (CRTE) before passing them to the language model~\cite{ICLR2026_5c1863f7}.
Its decoder separately uses one-dimensional rotary position embeddings (RoPE) to encode positions in the assembled input sequence~\cite{SU2024127063}.
We cyclically shift the encoded audio and video features before applying CRTE with the timestamps of their destination positions.
The audio--video shift uses the same target displacement for both modalities, but the realized displacements can differ after rounding to their respective temporal grids.

\paragraph{video-SALMONN~2+.}
video-SALMONN~2+ encodes temporal and spatial positions through multimodal RoPE in decoder attention~\cite{tang2025video}.
We cyclically shift audio-token embeddings and video temporal blocks at the decoder input while keeping the position IDs assigned to each input position fixed.
We convert the window duration $w$ into a number of video blocks using an estimated duration per block.
Within each window, we determine the audio-token shift from the video-block displacement using the estimated number of audio tokens per block.
Valid audio tokens are shifted as a single sequence, and padding tokens remain fixed.

\subsection{Baseline implementation}
\label{sec:supp_baselines}

We adapt the baseline methods to both backbones and keep their hyperparameters fixed across benchmarks.

\paragraph{AVCD.}
For AVCD~\cite{NEURIPS2025_5b5e70e7}, we use a contrastive strength of $0.5$, masking ratio of $0.5$, and entropy threshold of $0.6$ nats.
We use a plausibility parameter of $\beta=0.2$, following the official GitHub implementation.

\paragraph{ASD.}
We implement ASD~\cite{pmlr-v306-jung26c} following the paper's decoding formulation.
We use the sink dimensions $(458,2570)$ reported for video-SALMONN~2+, with an activation threshold of $25$ and persistence over at least eight decoder layers.
Since OmniVinci was not evaluated in the original ASD paper, we verified that these two dimensions account for 99.6\% of its decoder activations exceeding the threshold.
We rebalance attention with strength $0.6$ from decoder layer 6 onward and compute guidance statistics over layers 18--25 (zero-based indices).
Guidance is activated only when the unimodal sink attention share is at least $0.6$ and the text-attention mass is at most $0.5$.
We use a maximum guidance coefficient of $0.6$ and a momentum factor of $0.7$, initializing the previous coefficient to zero.

\begin{samepage}
In our first-token multiple-choice evaluation, the guidance coefficient is zero for 92.6--100\% of examples across the six settings.
The remaining nonzero corrections do not change the selected answers, resulting in identical predictions to Base.
\par
\end{samepage}

\paragraph{MAD and CAD.}
Both methods use predictions from audio--visual, audio-only, video-only, and text-only inputs.
A separate prompt estimates whether the question requires audio, video, or both.
MAD~\cite{Chung_2026_CVPR} uses the released weighting function with $\gamma=2.5$.
Our CAD~\cite{deng2026cad} reimplementation initializes the modality weights as in MAD and follows the paper's conflict and actionability calculations.
Following the paper, we use a conflict threshold of $\tau=0.05$ and a numerical stabilizer of $\epsilon=10^{-8}$.
Since the paper does not specify the transfer strength, we set $\lambda=1.0$ for both backbones across all benchmarks.

\subsection{Temporal perturbation strategies}
\label{sec:supp_ablation}

We keep the contrastive strength fixed across all variants in \cref{tab:abl_contrast,tab:abl_realisation,tab:abl_window_avut}, using $\alpha=5.0$ for OmniVinci and $\alpha=1.5$ for video-SALMONN~2+.

\paragraph{Positional offset.}
The positional-offset variant preserves feature content, token order, and sequence length while applying a global $+16$-second offset to the selected modality's temporal coordinates.
We choose the offset to approximate the 15-second target displacement of cyclic shift with $w=30$ seconds.
We round this target up to 16 seconds to match the two-second temporal block granularity of video-SALMONN~2+ under one-second frame sampling.
For OmniVinci, we add this offset to the selected modality's CRTE timestamps while keeping decoder position IDs unchanged.
For video-SALMONN~2+, we apply the corresponding offset to the selected modality's temporal position IDs, rounded to its temporal grid.

For video-SALMONN~2+, we also shift the position IDs of text tokens following the audio and video tokens by the same amount.
This prevents the text position IDs from overlapping with those of the shifted audio and video tokens.
We apply this text-position adjustment to all four conditions so that corresponding text tokens have identical position IDs.
Let $\mathbf z_P$ denote the logits obtained by adjusting only the text positions.
For OmniVinci, text positions remain unchanged, so $\mathbf z_P=\mathbf z$.
The logits $\mathbf z_{P,A}$, $\mathbf z_{P,V}$, and $\mathbf z_{P,AV}$ are obtained by additionally applying the offset to audio, video, or both, respectively.
The contrastive term is
\begin{equation}
    \Delta\mathbf z_{\rm pos}
    =\tfrac12(\mathbf z_P-\mathbf z_{P,A}
                       -\mathbf z_{P,V}+\mathbf z_{P,AV}).
\end{equation}
We add this term to the original-input logits as $\widetilde{\mathbf z}=\mathbf z+\alpha\Delta\mathbf z_{\rm pos}$.

\paragraph{Random permutation.}
We use the same window construction and position handling as cyclic shift (\cref{sec:supp_shifts}), with $w=30$ seconds.
For each window containing at least two video blocks, we sample a non-identity permutation of the blocks and their associated audio chunks.
We reuse this permutation across the audio shift, video shift, and audio--video shift conditions, preserving the pairing between video blocks and audio chunks in the audio--video shift.
Each block and chunk retains its internal token order.
For OmniVinci, audio features are grouped with the nearest video block in time, whereas video-SALMONN~2+ uses the audio tokens associated with each video block in the interleaved input.
For OmniVinci windows with fewer than two video blocks, we permute equal-duration audio chunks while leaving the video unchanged.
We use one permutation per window with a deterministic seed derived from the sample and window, without averaging over multiple seeds.

\subsection{Temporal error analysis protocol}
\label{sec:supp_temporal_errors}

We analyze incorrect-action selections on the original DAVE inputs using the main experimental settings.
Each of the 2,022 questions contributes three incorrect action options, giving 6,066 options in total.
Temporal distance is the absolute difference between the midpoints of the correct and incorrect action intervals.
We group distances into intervals of $[0,4), [4,8), \ldots, [16,20)$ seconds and a final interval of at least 20 seconds.
The \emph{none of the above} option remains available to the model, but we exclude it from the distance analysis because it has no associated action interval.

For each distance interval, $N$ counts incorrect action options from all questions, including those answered correctly.
For example, in the 12--16-second interval, OmniVinci selects 184 of 958 incorrect options with Base and 137 with MSCD, giving selection rates of 19.2\% and 14.3\%, respectively.

We also compute the mean temporal distance over questions for which the model selects an incorrect action, excluding correct answers and \emph{none of the above} selections.
This subset can differ between Base and MSCD.

\subsection{Audio-delay probe protocol}
\label{sec:supp_delay}

For each of the 2,022 DAVE Multimodal synchronization questions, we shift the full 16-kHz mono audio waveform by $\delta\in\{\pm4,\pm8,\pm16\}$ seconds.
For positive shifts, we prepend silence and trim the tail, whereas for negative shifts, we remove the beginning and append silence.
The video, clip duration, question, and answer-option order remain unchanged.

We use the annotated interval of the action onto which the cue sound was originally overlaid to determine the shifted target.
After shifting this interval by $\delta$ and clipping it to the video boundaries, we select the action with the largest temporal overlap.
Ties are resolved by selecting the earlier action.
If the same action occurs more than once, we use the answer option corresponding to the selected action interval.
We then exclude examples with no overlapping action or an unchanged target.
An example is included only if at least 70\% of the interval remaining after clipping overlaps the new action and at most 30\% overlaps the original action.
Target labels and subset selection are determined from the annotations, independently of the model's predictions.

We measure both selection rates on shifted inputs, pooling eligible examples from both directions at each shift magnitude.
$N$ counts eligible shifted examples rather than unique questions.
All five answer options, including \emph{none of the above}, remain available to the model.
For MSCD, we use the main experimental settings without probe-specific tuning.

\section{Additional Analyses}
\label{sec:supp_analyses}

\subsection{Video-only comparison on DAVE}
\label{sec:supp_video_only}

\Cref{tab:analysis_dave_audio_controls} compares Base and MSCD with a video-only condition that omits the audio input while keeping the video, question, and answer options unchanged.

  \par\addvspace{\textfloatsep}\noindent
  \begin{minipage}{\columnwidth}
    \renewenvironment{table}[1][]{}{\par}
    \captionsetup{type=table,hypcap=false}
    \begin{table}[t]
\centering
\small
\setlength{\tabcolsep}{4pt}
\caption{Accuracy (\%) on DAVE for Video only, Base, and MSCD. All conditions use the same 2,022 questions and original answer labels.}
\label{tab:analysis_dave_audio_controls}
\begin{adjustbox}{max width=\linewidth}
\renewcommand{\arraystretch}{1.15}
\setlength{\tabcolsep}{3pt}
\begin{tabular}{@{}l|cc@{}}
\toprule
\multicolumn{1}{@{}l!{\tableheadersingleseparator}}{Condition} & OmniVinci & video-SALMONN~2+ \\
\midrule
Video only & 41.6 & 40.8 \\
Base & 44.3 & 42.5 \\
MSCD & 49.8 & 45.5 \\
\bottomrule
\end{tabular}
\end{adjustbox}
\end{table}

  \end{minipage}\par\addvspace{\textfloatsep}

Even without audio, OmniVinci achieves 41.6\% accuracy and video-SALMONN~2+ achieves 40.8\%.
The accuracy gap from the video-only condition is modest for Base and larger for MSCD.
To investigate what may support the video-only performance, we examine the action durations associated with the answer choices.
The target action is the longest of the four annotated actions in 52.5\% of the questions.
Without audio, both models select the longest action in approximately 39\% of the questions.
This duration bias may partly explain the video-only accuracy.
Since action durations remain unchanged after an audio shift, this cue may also contribute to the model's preference for the original answer in the audio-shift probe (\cref{sec:obs_delay}).

\subsection{Effect of window duration on action confusion}
\label{sec:supp_temporal_windows}

We extend the temporal error analysis in \cref{sec:temporal_analysis_draft} (\cref{tab:analysis_dave_temporal_errors}) by comparing the main 30-second window with shorter 10- and 20-second windows.
Using the same DAVE questions and temporal-distance intervals, we report incorrect-action selection rates while keeping each backbone's contrastive strength fixed at its main value (\cref{tab:analysis_dave_temporal_errors_windows}).

For both backbones, 30-second windows yield lower selection rates than 10-second windows in every reported distance interval of at least 8 seconds.
For example, in the 16--20-second interval, OmniVinci's selection rate decreases from 22.7\% with 10-second windows to 15.6\% with 30-second windows.
However, the decrease is not always monotonic across window durations.
Below 8 seconds, shorter windows yield similar or slightly lower selection rates than 30-second windows.
Overall, larger windows tend to reduce confusion with temporally distant actions, while changes for nearby actions are comparatively small.

  \par\addvspace{\textfloatsep}\noindent
  \begin{minipage}{\columnwidth}
    \renewenvironment{table}[1][]{}{\par}
    \captionsetup{type=table,hypcap=false}
    \begin{table}[t]
\centering
\small
\setlength{\tabcolsep}{4pt}
\caption{Incorrect-action selection rates (\%) on DAVE for different window durations. The contrastive strength is fixed at the main value for each backbone. Subscripts show changes from Base in pp.}
\label{tab:analysis_dave_temporal_errors_windows}
\begin{adjustbox}{max width=\linewidth}
\renewcommand{\arraystretch}{1.15}
\setlength{\tabcolsep}{3pt}
\begin{tabular}{@{}l|cccc@{}}
\toprule
\multicolumn{1}{@{}l!{\tableheadersingleseparator}}{Distance (s)} & Base & 10 s & 20 s & 30 s \\
\midrule
\multicolumn{5}{@{}l}{\textit{OmniVinci}} \\
$[0,4)$ & 11.8 & 12.3\textsubscript{(+0.6)} & 13.4\textsubscript{(+1.7)} & 13.7\textsubscript{(+2.0)} \\
$[4,8)$ & 16.3 & 16.4\textsubscript{(+0.1)} & 15.7\textsubscript{(-0.6)} & 16.5\textsubscript{(+0.2)} \\
$[8,12)$ & 20.2 & 19.1\textsubscript{(-1.1)} & 17.7\textsubscript{(-2.5)} & 16.5\textsubscript{(-3.7)} \\
$[12,16)$ & 19.2 & 18.1\textsubscript{(-1.1)} & 17.2\textsubscript{(-2.0)} & 14.3\textsubscript{(-4.9)} \\
$[16,20)$ & 23.9 & 22.7\textsubscript{(-1.3)} & 19.1\textsubscript{(-4.8)} & 15.6\textsubscript{(-8.3)} \\
$\geq 20$ & 15.7 & 13.8\textsubscript{(-1.9)} & 14.6\textsubscript{(-1.1)} & 11.6\textsubscript{(-4.1)} \\
\midrule
\multicolumn{5}{@{}l}{\textit{video-SALMONN~2+}} \\
$[0,4)$ & 14.6 & 14.3\textsubscript{(-0.3)} & 14.0\textsubscript{(-0.6)} & 14.3\textsubscript{(-0.3)} \\
$[4,8)$ & 16.6 & 16.3\textsubscript{(-0.2)} & 16.5\textsubscript{(+0.0)} & 16.6\textsubscript{(+0.1)} \\
$[8,12)$ & 18.3 & 17.8\textsubscript{(-0.5)} & 17.1\textsubscript{(-1.2)} & 17.1\textsubscript{(-1.3)} \\
$[12,16)$ & 18.5 & 18.1\textsubscript{(-0.4)} & 16.2\textsubscript{(-2.3)} & 15.6\textsubscript{(-2.9)} \\
$[16,20)$ & 18.1 & 18.9\textsubscript{(+0.8)} & 15.4\textsubscript{(-2.8)} & 16.1\textsubscript{(-2.0)} \\
$\geq 20$ & 17.2 & 18.3\textsubscript{(+1.1)} & 17.5\textsubscript{(+0.4)} & 14.9\textsubscript{(-2.2)} \\
\bottomrule
\end{tabular}
\end{adjustbox}
\end{table}

  \end{minipage}\par\addvspace{\textfloatsep}

\begin{samepage}
\subsection{Audio-delay results by direction}
\label{sec:supp_delay_results}
  \par\addvspace{\textfloatsep}\noindent
  \begin{minipage}{\columnwidth}
    \renewenvironment{table}[1][]{}{\par}
    \captionsetup{type=table,hypcap=false}
    \begin{table}[t]
\centering
\small
\setlength{\tabcolsep}{4pt}
\caption{Selection rates (\%) on shifted DAVE inputs by audio-shift direction. Negative shifts advance the audio, while positive shifts delay it. $N$ counts the evaluated examples for each shift. Subscripts show changes from Base in pp.}
\label{tab:analysis_delay_directions}
\begin{adjustbox}{max width=\linewidth}
\renewcommand{\arraystretch}{1.15}
\begin{tabular}{@{}cc|cc|cc@{}}
\toprule
\multirow{2}{*}{Delay (s)} & \multirow{2}{*}{$N$} & \multicolumn{2}{c|}{\shortstack{Shifted correct\\answer (\%) $\uparrow$}} & \multicolumn{2}{c}{\shortstack{Original correct\\answer (\%)}} \\
\cmidrule(lr){3-4}\cmidrule(lr){5-6}
 & & Base & MSCD & Base & MSCD \\
\midrule
$-16$ & 580 & 24.8 & \textbf{34.8}\textsubscript{\textcolor{mscdincrease}{(+10.0)}} & 37.6 & 24.3\textsubscript{(-13.3)} \\
$-8$ & 852 & 19.5 & \textbf{28.1}\textsubscript{\textcolor{mscdincrease}{(+8.6)}} & 41.7 & 42.4\textsubscript{(+0.7)} \\
$-4$ & 329 & 16.1 & \textbf{20.1}\textsubscript{\textcolor{mscdincrease}{(+4.0)}} & 39.8 & 41.6\textsubscript{(+1.8)} \\
\midrule
$+4$ & 274 & 30.7 & \textbf{32.5}\textsubscript{\textcolor{mscdincrease}{(+1.8)}} & 30.7 & 31.4\textsubscript{(+0.7)} \\
$+8$ & 605 & \textbf{32.7} & 31.7\textsubscript{\textcolor{mscddecrease}{(-1.0)}} & 32.7 & 31.2\textsubscript{(-1.5)} \\
$+16$ & 358 & \textbf{36.9} & \textbf{36.9}\textsubscript{(+0.0)} & 29.1 & 31.0\textsubscript{(+2.0)} \\
\bottomrule
\end{tabular}
\end{adjustbox}
\end{table}

  \end{minipage}\par\addvspace{\textfloatsep}

\end{samepage}

\Cref{tab:analysis_delay_directions} reports the results in \cref{tab:analysis_delay_mscd} separately for advanced and delayed audio.
The pooled gains arise primarily from examples where the audio is advanced.
In this condition, MSCD increases the selection rate of the shifted correct answer at all three tested magnitudes, with the largest gain of 10.0~pp at $-16$ seconds.
For delayed audio, changes in this selection rate range from $-1.0$ to $+1.8$~pp.
Because the eligible subsets differ across directions and magnitudes, the observed differences cannot be attributed to shift direction or magnitude alone.

\subsection{Window duration and video duration on AVUT}
\label{sec:supp_window_duration}

To examine how window duration affects performance on videos of different lengths, we evaluate six window settings on AVUT (\cref{tab:analysis_window_duration_avut}).
All settings use the same 1,120 questions and the backbone-specific contrastive strengths used in the main results.
When a window spans the entire clip, the perturbation is identical to the Whole setting.

  \par\addvspace{\textfloatsep}\noindent
  \begin{minipage}{\columnwidth}
    \renewenvironment{table}[1][]{}{\par}
    \captionsetup{type=table,hypcap=false}
    \begin{table}[t]
\centering
\small
\setlength{\tabcolsep}{4pt}
\caption{Accuracy (\%) on AVUT by window and video duration. Whole uses a single window spanning the entire clip. QA denotes the number of questions in each group. Subscripts show changes from Base in pp.}
\label{tab:analysis_window_duration_avut}
\begin{adjustbox}{max width=\linewidth}
\renewcommand{\arraystretch}{1.15}
\setlength{\tabcolsep}{2pt}
\begin{tabular}{@{}l|cccc|c@{}}
\toprule
\multirow{2}{*}{Window} & \multicolumn{4}{c|}{Video duration (s)} & \multirow{2}{*}{All} \\
\cmidrule(lr){2-5}
 & $\leq30$ & $(30,60]$ & $(60,120]$ & $>120$ & \\
\midrule
QA & 264 & 406 & 379 & 71 & 1,120 \\
\midrule
\multicolumn{6}{@{}l}{\textit{OmniVinci}} \\
Base & 63.6 & 68.7 & 66.5 & 64.8 & 66.5 \\
\midrule
10 s & 66.3\textsubscript{\textcolor{mscdincrease}{(+2.7)}} & 65.8\textsubscript{\textcolor{mscddecrease}{(-3.0)}} & 66.8\textsubscript{\textcolor{mscdincrease}{(+0.3)}} & 64.8\textsubscript{(+0.0)} & 66.2\textsubscript{\textcolor{mscddecrease}{(-0.4)}} \\
20 s & 66.7\textsubscript{\textcolor{mscdincrease}{(+3.0)}} & 66.3\textsubscript{\textcolor{mscddecrease}{(-2.5)}} & 66.8\textsubscript{\textcolor{mscdincrease}{(+0.3)}} & 57.7\textsubscript{\textcolor{mscddecrease}{(-7.0)}} & 66.0\textsubscript{\textcolor{mscddecrease}{(-0.5)}} \\
30 s & 67.8\textsubscript{\textcolor{mscdincrease}{(+4.2)}} & 69.7\textsubscript{\textcolor{mscdincrease}{(+1.0)}} & 66.8\textsubscript{\textcolor{mscdincrease}{(+0.3)}} & 60.6\textsubscript{\textcolor{mscddecrease}{(-4.2)}} & 67.7\textsubscript{\textcolor{mscdincrease}{(+1.2)}} \\
60 s & 67.8\textsubscript{\textcolor{mscdincrease}{(+4.2)}} & 67.0\textsubscript{\textcolor{mscddecrease}{(-1.7)}} & 65.4\textsubscript{\textcolor{mscddecrease}{(-1.1)}} & 64.8\textsubscript{(+0.0)} & 66.5\textsubscript{(+0.0)} \\
120 s & 67.8\textsubscript{\textcolor{mscdincrease}{(+4.2)}} & 67.0\textsubscript{\textcolor{mscddecrease}{(-1.7)}} & 64.6\textsubscript{\textcolor{mscddecrease}{(-1.8)}} & 69.0\textsubscript{\textcolor{mscdincrease}{(+4.2)}} & 66.5\textsubscript{(+0.0)} \\
Whole & 67.8\textsubscript{\textcolor{mscdincrease}{(+4.2)}} & 67.0\textsubscript{\textcolor{mscddecrease}{(-1.7)}} & 64.6\textsubscript{\textcolor{mscddecrease}{(-1.8)}} & 64.8\textsubscript{(+0.0)} & 66.2\textsubscript{\textcolor{mscddecrease}{(-0.3)}} \\
\midrule
\multicolumn{6}{@{}l}{\textit{video-SALMONN 2+}} \\
Base & 54.5 & 62.1 & 65.7 & 66.2 & 61.8 \\
\midrule
10 s & 57.2\textsubscript{\textcolor{mscdincrease}{(+2.7)}} & 63.8\textsubscript{\textcolor{mscdincrease}{(+1.7)}} & 66.0\textsubscript{\textcolor{mscdincrease}{(+0.3)}} & 62.0\textsubscript{\textcolor{mscddecrease}{(-4.2)}} & 62.9\textsubscript{\textcolor{mscdincrease}{(+1.1)}} \\
20 s & 59.1\textsubscript{\textcolor{mscdincrease}{(+4.5)}} & 61.8\textsubscript{\textcolor{mscddecrease}{(-0.2)}} & 64.6\textsubscript{\textcolor{mscddecrease}{(-1.1)}} & 64.8\textsubscript{\textcolor{mscddecrease}{(-1.4)}} & 62.3\textsubscript{\textcolor{mscdincrease}{(+0.5)}} \\
30 s & 59.1\textsubscript{\textcolor{mscdincrease}{(+4.5)}} & 61.3\textsubscript{\textcolor{mscddecrease}{(-0.7)}} & 65.4\textsubscript{\textcolor{mscddecrease}{(-0.3)}} & 64.8\textsubscript{\textcolor{mscddecrease}{(-1.4)}} & 62.4\textsubscript{\textcolor{mscdincrease}{(+0.6)}} \\
60 s & 59.1\textsubscript{\textcolor{mscdincrease}{(+4.5)}} & 62.6\textsubscript{\textcolor{mscdincrease}{(+0.5)}} & 64.6\textsubscript{\textcolor{mscddecrease}{(-1.1)}} & 59.2\textsubscript{\textcolor{mscddecrease}{(-7.0)}} & 62.2\textsubscript{\textcolor{mscdincrease}{(+0.4)}} \\
120 s & 59.1\textsubscript{\textcolor{mscdincrease}{(+4.5)}} & 62.6\textsubscript{\textcolor{mscdincrease}{(+0.5)}} & 64.4\textsubscript{\textcolor{mscddecrease}{(-1.3)}} & 64.8\textsubscript{\textcolor{mscddecrease}{(-1.4)}} & 62.5\textsubscript{\textcolor{mscdincrease}{(+0.7)}} \\
Whole & 59.1\textsubscript{\textcolor{mscdincrease}{(+4.5)}} & 62.6\textsubscript{\textcolor{mscdincrease}{(+0.5)}} & 64.4\textsubscript{\textcolor{mscddecrease}{(-1.3)}} & 62.0\textsubscript{\textcolor{mscddecrease}{(-4.2)}} & 62.3\textsubscript{\textcolor{mscdincrease}{(+0.5)}} \\
\bottomrule
\end{tabular}
\end{adjustbox}
\end{table}

  \end{minipage}\par\addvspace{\textfloatsep}

MSCD outperforms Base on videos up to 30 seconds under every tested window setting.
With the main 30-second window, accuracy increases by 4.2~pp for OmniVinci and 4.5~pp for video-SALMONN~2+ in this group.
Overall, only the 30-second window improves accuracy over Base for OmniVinci, whereas video-SALMONN~2+ outperforms Base under all six settings.

Gains are less consistent on longer videos.
On videos longer than 120 seconds, only the 120-second window improves OmniVinci's accuracy over Base, whereas all tested settings remain below Base for video-SALMONN~2+.
This group contains only 71 questions, so these results should be interpreted with caution.

\begin{samepage}
\subsection{Video duration analysis on Daily-Omni}
\label{sec:supp_dailyomni_duration}
  \par\addvspace{\textfloatsep}\noindent
  \begin{minipage}{\columnwidth}
    \renewenvironment{table}[1][]{}{\par}
    \captionsetup{type=table,hypcap=false}
    \begin{table}[t]
\centering
\small
\setlength{\tabcolsep}{4pt}
\caption{Accuracy (\%) on Daily-Omni by video duration using the main experimental settings. Subscripts show changes from Base in pp.}
\label{tab:analysis_duration_dailyomni}
\begin{adjustbox}{max width=\linewidth}
\renewcommand{\arraystretch}{1.15}
\begin{tabular}{@{}l|cc|cc@{}}
\toprule
\multirow{2}{*}{Method} & \multicolumn{2}{c}{OmniVinci} & \multicolumn{2}{c}{video-SALMONN~2+} \\
\cmidrule(lr){2-3}\cmidrule(lr){4-5}
 & 30 s & 60 s & 30 s & 60 s \\
\midrule
Base & 62.9 & 59.1 & 62.3 & 60.7 \\
\textbf{Ours} & \textbf{65.1}\textsubscript{\textcolor{mscdincrease}{(+2.2)}} & \textbf{61.3}\textsubscript{\textcolor{mscdincrease}{(+2.2)}} & \textbf{64.9}\textsubscript{\textcolor{mscdincrease}{(+2.6)}} & \textbf{63.8}\textsubscript{\textcolor{mscdincrease}{(+3.1)}} \\
\bottomrule
\end{tabular}
\end{adjustbox}
\end{table}

  \end{minipage}\par\addvspace{\textfloatsep}

\end{samepage}

To examine whether the concentration of gains in shorter AVUT videos also appears on Daily-Omni, we compare its 30- and 60-second groups (\cref{tab:analysis_duration_dailyomni}).
These groups contain 647 and 550 questions, respectively.
MSCD improves accuracy over Base in both groups, with comparable or larger gains on 60-second videos than on 30-second videos.
This contrasts with the pattern observed on AVUT.

\subsection{Inference time comparison}
\label{sec:supp_computational_cost}

We measure first-token inference time on 100 DAVE questions using a single NVIDIA TITAN RTX GPU with FP16 precision.
Each method is evaluated separately after two warm-up questions.
We execute forward passes sequentially with a batch size of one and synchronize the GPU for timing.
For AVCD, masked forward passes are excluded when its entropy gate is closed.

  \par\addvspace{\textfloatsep}\noindent
  \begin{minipage}{\columnwidth}
    \renewenvironment{table}[1][]{}{\par}
    \captionsetup{type=table,hypcap=false}
    \begin{table}[t]
\centering
\small
\setlength{\tabcolsep}{4pt}
\caption{Median per-question inference time, including audio and video encoding. Relative latency is the median of the per-question ratios to Base.}
\label{tab:analysis_decoding_latency}
\begin{adjustbox}{max width=\linewidth}
\renewcommand{\arraystretch}{1.15}
\setlength{\tabcolsep}{3pt}
\begin{tabular}{@{}lc|cc|cc@{}}
\toprule
 & & \multicolumn{2}{c|}{OmniVinci} & \multicolumn{2}{c}{video-SALMONN~2+} \\
\cmidrule(lr){3-4}\cmidrule(l){5-6}
Method & \shortstack{LLM\\forwards} & Time (s) & $\times$ Base & Time (s) & $\times$ Base \\
\midrule
Base & 1 & 2.22 & 1.00 & 0.60 & 1.00 \\
AVCD & 1 or 4 & 8.16 & 3.77 & 1.66 & 2.93 \\
ASD & 2 & 5.00 & 2.26 & 1.13 & 1.95 \\
MAD & 5 & 3.91 & 1.77 & 1.17 & 1.99 \\
CAD & 5 & 3.85 & 1.70 & 1.17 & 1.99 \\
MSCD & 4 & 4.26 & 1.91 & 1.42 & 2.38 \\
\bottomrule
\end{tabular}
\end{adjustbox}
\end{table}

  \end{minipage}\par\addvspace{\textfloatsep}

MSCD takes approximately $1.9\times$ and $2.4\times$ the time of Base on OmniVinci and video-SALMONN~2+, respectively, within the range of the evaluated decoding baselines (\cref{tab:analysis_decoding_latency}).
Although MSCD requires four LLM forward passes, all four conditions share the encoder outputs.
MAD and CAD use shorter unimodal or text-only inputs for three of their five passes, whereas our AVCD and ASD implementations compute attention weights without optimized attention kernels.

\section{Qualitative Examples}
\label{sec:supp_qualitative}
\Cref{fig:supp_qualitative_tca} compares caption excerpts generated by Base and MSCD using video-SALMONN~2+ on four selected TCA-Bench~\cite{10.1007/978-3-032-37490-5_4} videos.
In these examples, Base attributes speech to the wrong person, whereas MSCD associates it with the correct speaker.

Base and MSCD receive the same prompt requesting speech, other sounds, their sources, and accompanying visual events in chronological order.
Both methods use greedy decoding with a maximum of 192 generated tokens.
For MSCD, we use $\alpha=1.5$, $w=30$ seconds, and an adaptive plausibility constraint~\cite{li-etal-2023-contrastive,Leng_2024_CVPR} with $\beta=0.1$.

\twocolumn[{
\begin{minipage}{\textwidth}
\centering
\includegraphics[width=\linewidth]{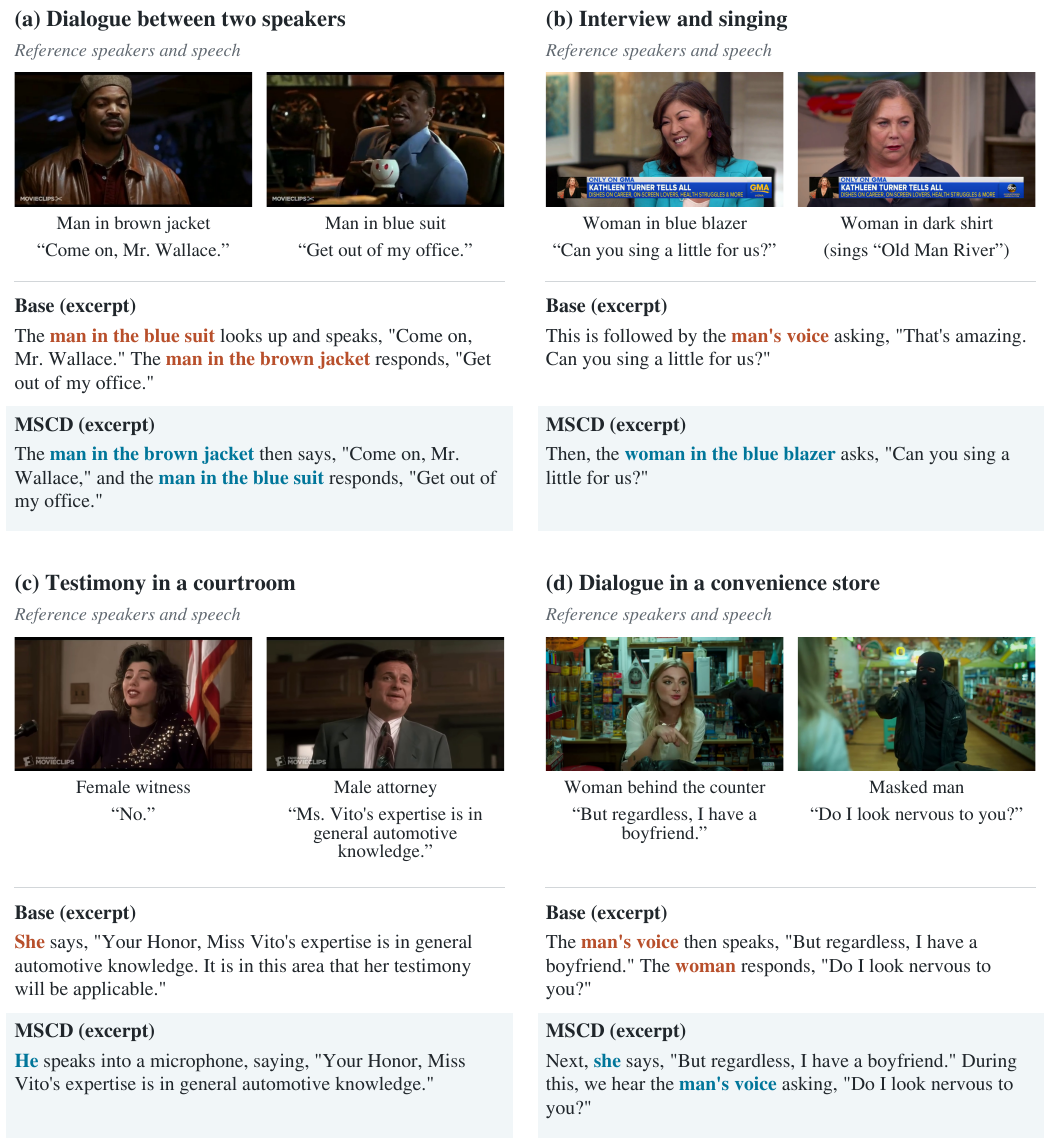}
\captionof{figure}{Qualitative caption excerpts from video-SALMONN~2+ on TCA-Bench.
Representative video frames are paired with speaker descriptions and speech from the dataset annotations.
Orange and blue highlight incorrect speaker assignments in Base and the corresponding corrections in MSCD, respectively.
The excerpts are reproduced verbatim.}
\label{fig:supp_qualitative_tca}
\end{minipage}
}]

\end{document}